\documentclass{elsart}

\usepackage{amsmath}
\usepackage{graphicx}
\usepackage{stmaryrd}
\usepackage{amssymb}
\usepackage{color}
\usepackage{multirow}
\usepackage{lineno}
\usepackage{algorithm}
\usepackage{algorithmic}
\usepackage{lscape} 
\usepackage{subcaption}
\usepackage{stmaryrd}
\usepackage{todonotes}

\usepackage{lscape}

\usepackage{url}

\usepackage{caption}
\usepackage[toc]{appendix}

\begin{document}

\begin{frontmatter}

\title{A massively parallel evolutionary algorithm for the partial Latin square extension problem}


\author{Olivier Goudet} \and
\ead{olivier.goudet@univ-angers.fr}
\author{Jin-Kao Hao\corauthref{cor}}
\ead{jin-kao.hao@univ-angers.fr}
\corauth[cor]{Corresponding author.}
\address{LERIA, University of Angers, 2 Boulevard Lavoisier, 49045 Angers, France}



\begin{abstract}
The partial Latin square extension problem is to fill as many as possible empty cells of a partially filled Latin square. This problem is a useful model for a wide range of applications in diverse domains. This paper presents the first massively parallel evolutionary algorithm algorithm for this computationally challenging problem based on a transformation of the problem to partial graph coloring. The algorithm features the following original elements. Based on a very large population (with more than $10^4$ individuals) and modern graphical processing units, the algorithm performs many local searches in parallel to ensure an intensive exploitation of the search space. The algorithm employs a dedicated crossover with a specific parent matching strategy to create a large number of diversified and information-preserving offspring at each generation. Extensive experiments on 1800 benchmark instances show a high competitiveness of the algorithm compared to the current best performing methods. Competitive results are also reported on the related Latin square completion problem. Analyses are performed to shed lights on the roles of the main algorithmic components. The code of the algorithm will be made publicly available. \\ 

\end{abstract}

\end{frontmatter}


\section{Introduction}\label{Introduction}

\begin{figure}[h]
    \centering
    \includegraphics[width=0.65\textwidth]{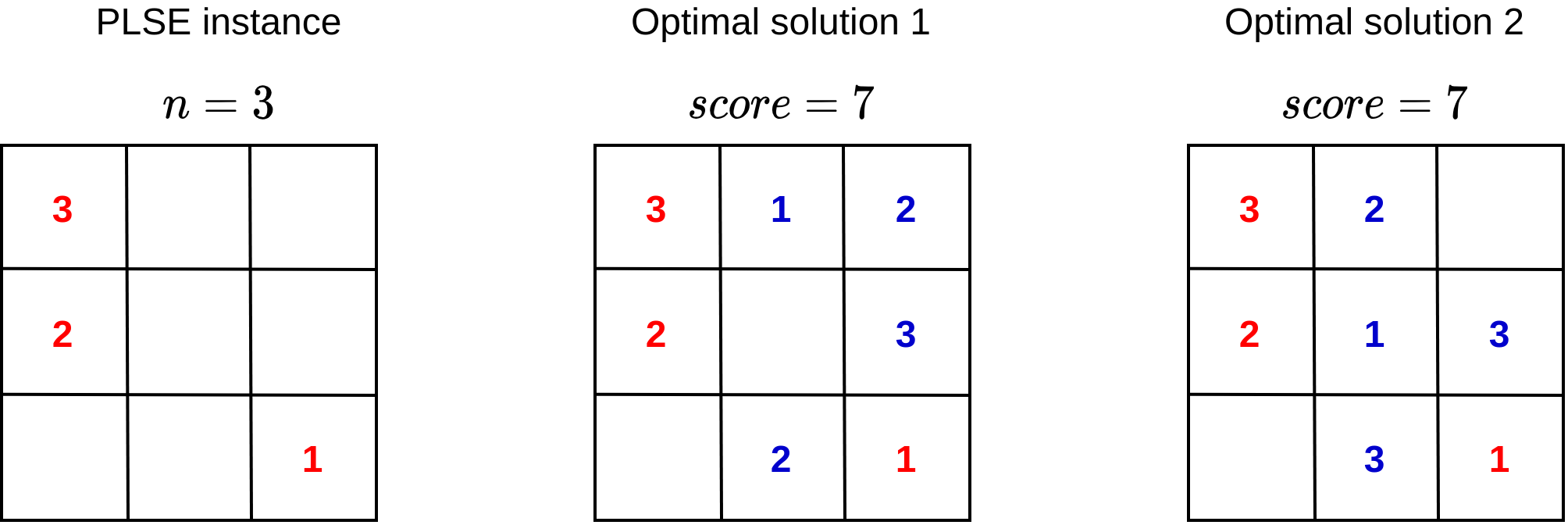}
    \caption{Example of a PLSE instance with $n=3$.  }
    \label{fig:PLSE_1} 
\end{figure}

A Latin square $\mathcal{L}$ of order $n$, also called a Quasigroup, is an $n \times n$ grid filled with $n$ distinct symbols $\{1,\dots,n\}$, where each symbol appears exactly once in each row and column of the grid (Latin square condition). A partial Latin square of order $n$ verifies that each cell is either empty or contains one of the $n$ symbols, and each symbol occurs at most once in any row or column. Given a partial Latin square, the partial Latin square extension (PLSE) problem is to fill as many as possible the empty cells. The Latin square completion (LSC) problem (also called Quasigroup completion problem) is the decision version that determines whether it is possible to fill all the empty cells of a partial Latin square. Figure \ref{fig:PLSE_1} shows a PLSE instance with $n=3$. Numbers in red correspond to filled cells. Two different optimal solutions with a score of 7 are displayed (it is impossible to complete the grid). 

\begin{figure}[h]
    \centering
    \includegraphics[width=0.65\textwidth]{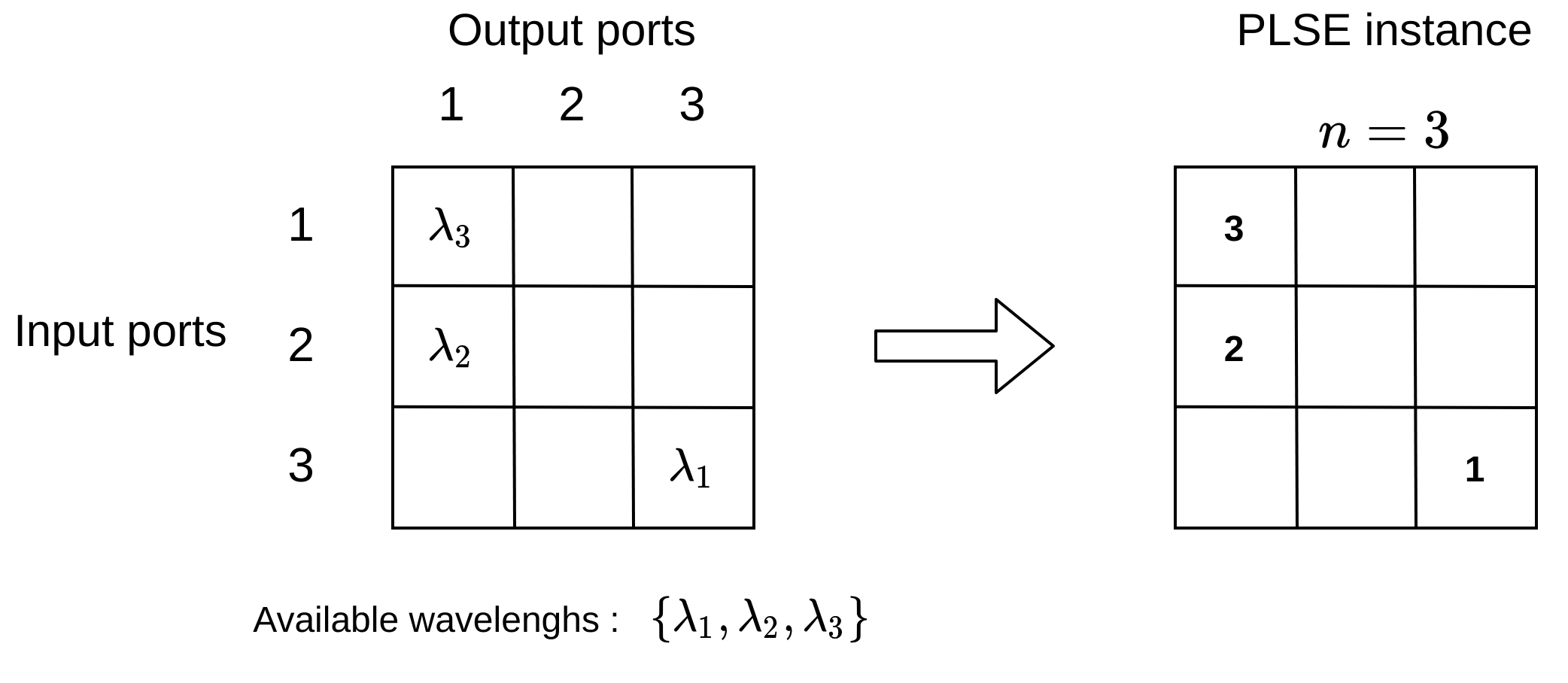}
    \caption{ Example of the optical router extension problem with $n=3$ converted to the partial Latin square extension problem.}
    \label{fig:optical_router}
\end{figure}

Latin square problems naturally appear in numerous applications, such as scheduling, error correcting codes, as well as experimental and combinatorial design \cite{LatinSquaresAppli2015,jakobovic2021toward}. For instance, a typical application of the PLSE is the design of optical router systems \cite{barry1993Latin}. Optical routers are connected by optical links supporting a number of wavelengths. Each router has $n$ input and $n$ output links, and is capable of switching wavelengths to avoid conflicts in optical links. As displayed on Figure \ref{fig:optical_router}, the connections between input and output links can be modeled by a $n \times n$ array, where the $n$ rows and $n$ columns correspond to the inputs and the $n$ outputs respectively. Each element at the position $(i,j)$ can be filled by a number in $\{1,\dots, n\}$ indicating a specific   wavelength used for the connection between input port $i$ and output port $j$. In order to avoid wavelength conflicts, it is mandatory that each input or output port does not use the same frequency for more than one communication. Empty connections between input and output are indicated with empty cells. Given a router with existing connections between inputs and outputs, adding as many as new connections without introducing conflicts in the router is to solve the associated partial Latin square extension problem.

The LSC is known to be NP-complete \cite{colbourn1984complexity}. As the result, both the decision problem (LCS) and the optimization problem (PLSE) are computationally challenging in the general case. Due to their importance, Latin square problems have been studied from a wide variety of perspectives in different fields. 

In algebra, the multiplication table of a finite quasigroup corresponds to a Latin square \cite{evans1960embedding}. As such, Latin squares have been studied as a mathematical object and various  properties were established \cite{bennett1989quasigroup,artamonov2016characterization,artamonov2013Latin}.

The LSC can be expressed as an integer program with $n^3$ Boolean variables $x_{i,j,k}$, where $x_{i,j,k} = 1$ indicates that the cell in position $(i,j)$ receives the symbol $k \in \{1,\dots, n \}$. With this formulation and using integer programming solvers, optimal results were reported for small instances in \cite{gomes2002completing}. The authors also investigated two other exact methods based on constraint programming (CP) and SAT technologies. A systematic comparison of SAT and CP models was presented in \cite{AnsoteguiVDFM04}. An approximation algorithm was proposed based on a packing integer programming formulation in \cite{gomes2004improved}. 

In terms of practically solving the PLSE, a notable work was presented in \cite{haraguchi2016iterated}. In this work, a partial Latin square was represented by means of an orthogonal array, with a set of triples in $[n]^3$, such that each element $(v_1,v_2,v_3)$ of this set indicates that the symbol $v_3$ is assigned to $(v_1,v_2)$. The Hamming distance between each pair of elements of this set is then computed. If the distance between each pair of triples of this set is at least two, this set corresponds to a partial Latin square. Based on this representation, the author proposed several iterated local search algorithms which aim to extend the current set of triples without adding conflicts. To assess the practical performance of these iterated local search algorithms, the author introduced a set of 1800 instances for the PLSE and another set of 1800 instances for the LSC with various characteristics (cf. Section \ref{Benchmark} and Appendix \ref{sec:results_LSC}). Computational results showed that the iterated local search algorithms perform extremely well by outperforming previous methods including integer programming, constraint programming as well as their hybridized approach. 

The problem of extending a partial Latin square can also be studied under the view of (partial) graph coloring \cite{lewis2015guide}. Indeed, a partial Latin square of order $n$ can be transformed to a graph such that each vertex (or node) corresponds to a cell of the grid (there are thus $n^2$ vertices), and an edge exists between two vertices corresponding to two cells of the same row or column (there are thus $n^2(n-1)$ edges). The vertex of a pre-filled cell with a symbol $k$ receives color $k \in \{1,\dots, n\}$. Empty cells are left uncolored. The PLSE is to color as many uncolored vertices as possibles such that two colored vertices do not share the same color. Based on this observation, the authors of \cite{jin2019solving} proposed a powerful memetic algorithm (MMCOL) for the Latin square completion problem and solved all the 1800 LSC instances introduced in \cite{haraguchi2016iterated} as well as all the 19 traditional LSC instances in the literature \cite{gomes2002completing}. With some slight adaptations of their algorithm, they also reported excellent results on the 1800 PLSE instances of \cite{haraguchi2016iterated}.

To sum, the two most recent studies on the PLSE \cite{haraguchi2016iterated} and the LSC \cite{jin2019solving} significantly contributed to the practical solving of these two challenging problems. In particular, all the existing LSC benchmark instances have been solved thanks to the algorithm presented in \cite{jin2019solving}. On the contrary, this is not the case for the PLSE and there is still room for improvement in terms of better solving the PLSE instances. In fact, for almost half of the 1800 benchmark instances, their optimal solutions are still unknown and only lower bounds were reported. 

This work is motivated by this observation, and aims to advance the state-of-the-art of solving the PLSE by establishing record-breaking lower bounds for the unsolved PLSE instances. For this purpose, we introduce the first massively parallel evolutionary algorithm algorithm that fully takes advantage of the GPU architecture to parallelize all critical search components. We summarize the contributions of this work as follows.

From the perspective of algorithm design, the proposed algorithm relies on a very large population $P$ ($|P|>10^4$) that enables massively parallel local optimization and offspring generation on the GPU architecture. This is in sharp contrast to the typical use of a small population $P$ (typically $|P|<10^2$) and sequential computations of many memetic algorithms including the MMCOL algorithm \cite{banos2010memetic,ruiz2019parallel,jin2019solving}. The algorithm features several complementary and original search components including a parametrized asymmetric uniform crossover and an effective local search. The crossover uses a probability to control the inherited information from the parents according to a distance metric and a specific parent matching strategy to create a large number of diversified and information-preserving offspring. The local search utilizes a two-phase approach to effectively explore an enlarged search space. The algorithm is further reinforced by a parallel distance calculation procedure that enables a fast population updating.

From the perspective of computational performance, we demonstrate a high competitiveness of our algorithm on the 1800 PLSE benchmark instances from \cite{haraguchi2016iterated}. We report many improved best lower bounds for large and difficult instances, including 25 record optimal solutions. We also test the algorithm on the related LSC and show that the algorithm is able to solve all the existing benchmark instances (1800 from \cite{haraguchi2016iterated} and 19 from \cite{gomes2002completing}).

Finally, we contribute to the understanding of the population size, the crossover  and the parent matching strategy for a large population. In particular, we show that the random parent matching strategy which is typically employed in many memetic algorithms (e.g., \cite{lu2010memetic,jin2019solving}) is no more suitable in the context of a large population and can be beneficially replaced by a neighborhood matching strategy for a better efficiency. 

In the rest of the paper, we present the solution approach and the proposed algorithm (Sections \ref{sec:PLSE} and \ref{sec:MPMA}), experimental results and comparisons with the state-of-the-art methods (Sections \ref{sec:experiments}), followed by analyses of key algorithmic components and conclusions (Sections \ref{sec:key_components} and \ref{sec:conclusion}).

\section{Partial Latin square extension as graph coloring} \label{sec:PLSE}

This section illustrates how the partial Latin square extension problem can be considered as a graph coloring problem. This approach was first used in \cite{jin2019solving} with a great success to solve the related Latin square completion problem. However, two specific and significant features of the partial Latin square extension problem were ignored until now. We discuss them at the end of this section, which also provide additional motivations for this work.

\subsection{Partial Latin square extension to Latin square graph}\label{Latin square graph}

Let $\mathcal{L}$ be a Latin square of order $n$ composed of $n \times n$ cells. $\mathcal{L}$ can be transformed into a graph $G=(V,E)$, called Latin square graph, with the vertex set $V = \{\{1, \dots, n\} \times \{1, \dots, n \}\}$ of size $|V| = n^2$ and the edge set $E$ of size $|E| = n^2(n-1)$ where $\{u,v\} \in E$ if and only if $u$ and $v$ are two vertices representing two cells of the same row or same column of $\mathcal{L}$ \cite{lewis2015guide,jin2019solving}. Then solving the PLSE can be reached by finding a partial legal $n$-coloring (also called list-coloring \cite{lewis2015guide}) of the graph $G$ by using the colors in $\{1,...,n\}$ while maximizing the number of colored vertices (or equivalently minimizing the number of uncolored vertices). 

Let $D(v)$ denote the color domain of vertex $v$ (i.e., the set of colors that can be used to color $v$). If $v$ corresponds to a cell pre-filled with symbol $k$ ($k\in \{1,...,n\}$), $D(v)=\{k\}$. If $v$ corresponds to an empty cell, $v$ can receive a color in $\{1,...,n\}$ or remain uncolored, indicated with the color $0$. In other words, $D(v) = \{0,1,...,n\}$ for any vertex $v$ representing an empty cell. Then a (partial) legal $n$-coloring  of the associated Latin square graph $G$ is a function $S:V \rightarrow \{D(v_1),\dots,D(v_{|V|})\}$ such that for any pair of vertices $u$ and $v$, if $S(u) \neq 0$, $S(v) \neq 0$, and they are linked by an edge ($\{u,v\} \in E$), then their colors $S(u)$ and $S(v)$ must be different ($S(u) \neq S(v)$). Note that a vertex receiving color 0 indicates an uncolored vertex.

A legal solution of the PLSE can also be seen as a partition of $V$ into $n$ disjoint stable sets $V_1$, $V_2$, ..., $V_n$ and a set $V_0 = V \backslash \cup_{i=1}^n V_i$,  such that $V_i$ is the set of vertices receiving color $i$. A set $V_i$ ($i=1,\dots,n$) is a stable set if $\forall (u,v) \in V_i, \{u,v\} \notin E$. A stable set is also called a color class.

Let $S= \{V_0, V_1$, $V_2$, ..., $V_n\} $ be a partition of the vertex set $V$, the objective of the partial Latin square extension problem (PLSE) in terms of the list-coloring problem can be stated as follows:
 \begin{equation}
     \text{minimize} \ f(S) = |V_0|
     \label{eq:fitness}
 \end{equation}
 \begin{equation}
     \text{subject to} \ \forall u, v \in V_i, \{u,v\} \notin E, i=1,2 \dots, n
     \label{eq:constraints}
 \end{equation}

\noindent where the objective (\ref{eq:fitness}) is to minimize the cardinality of the set $V_0$ (number of uncolored vertices) and the constraints (\ref{eq:constraints}) ensure that the partition $\{ V_0, V_1$, $V_2$, \dots, $V_n\}$ is a legal but potentially partial $n$-coloring. 
Notice that this formulation of the partial Latin square extension problem can also be used to solve the Latin square completion problem (LSC), for which a legal solution $S$ with $f(S)=0$ is sought.


The constraints (\ref{eq:constraints}) can be reformulated with a constraint function $c$ which simply counts the number of conflicts in $S$:
\begin{equation} \label{f}
    c(S) = \sum_{\{u,v\} \in E} \delta_{uv},
\end{equation}
\noindent where
\begin{equation}
    \delta_{uv} = \begin{cases}
      1 & \text{if $u \in V_i$, $v \in V_j$, $i=j$ and $i\neq 0$}\\
      0 & \text{otherwise}.
    \end{cases}  
\end{equation}

If $\delta_{uv} = 1$, $u$ and $v$ are two conflicting vertices (i.e., they receive the same colors while they are adjacent in the graph). Clearly, a coloring $S$ with $c(S) = 0$ corresponds to a legal $n$-coloring.

\begin{figure}[h]
    \centering
    \includegraphics[width=0.65\textwidth]{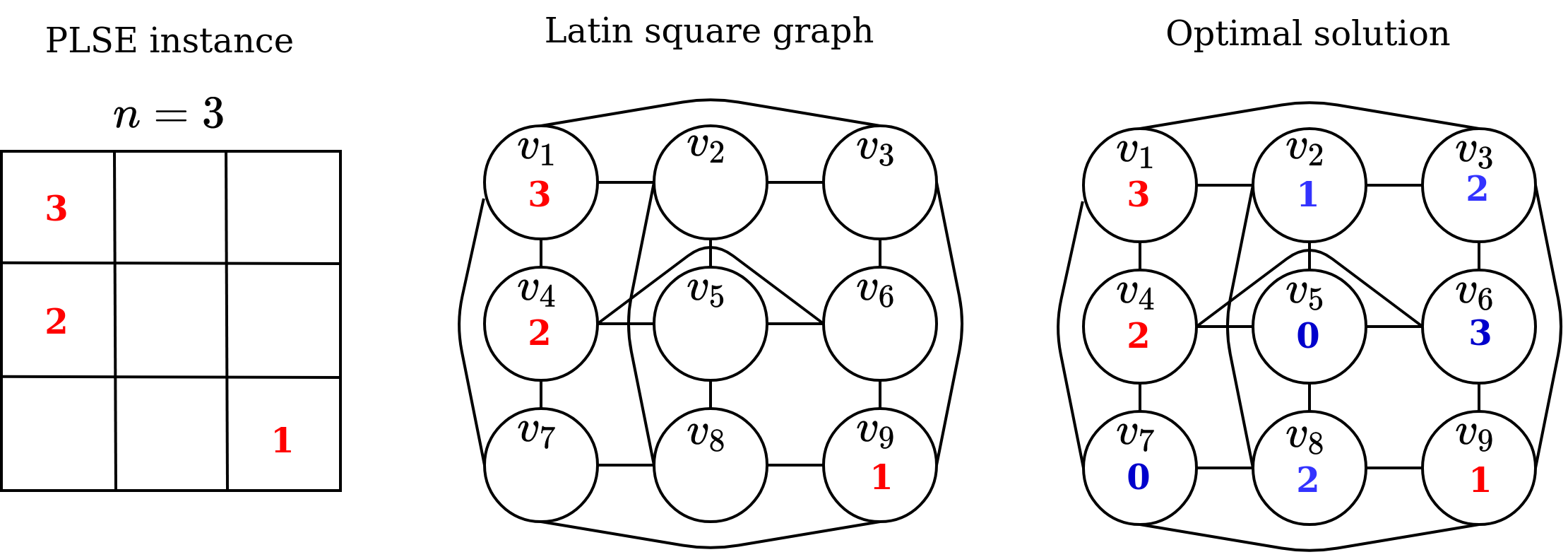}
    \caption{\label{fig:PLSE_2} Example of converting a partial Latin square extension instance (left) to a Latin square graph (middle) and an optimal partial coloring with two uncolored vertices (color 0) (right). }
\end{figure}

Figure \ref{fig:PLSE_2} shows a PLSE instance (left), its Latin square graph (middle) and a legal partial coloring of the Latin square graph with two uncolored vertices (right).

\subsection{Preprocessing of the Latin square graph}\label{preprocess_Latin_square}

As indicated in \cite{jin2019solving}, a preprocessing procedure can be applied to reduce a Latin square graph by removing the colored vertices (i.e., the filled cells). Indeed, if a vertex $v$ of  the graph represents a cell pre-filled with symbol $k\in \{1,...,n\}$, the vertex definitively receives this unique color $k$ and can be removed from the graph. Moreover, since the color $k$ cannot be assigned to any vertex $u$ adjacent to $v$ (i.e., $\{u,v\}\in E$), this color can thus be safely removed from the color domain $D(u)$ (in order to respect the Latin square condition).
  
Nevertheless, during the preprocessing, if the color domain of a vertex $u$ becomes the singleton $D(u) = \{0\}$, it means that the corresponding cell cannot be filled. This cell remains definitively unfilled and the vertex $u$ is removed from the graph. If one denotes by $l$ the number of cells impossible to fill after this preprocessing phase, $n^2 - l$ defines an upper bound of the optimal value (score) of the given PLSE instance. For the special case of $l=1$, a better upper bound is in fact $n^2 - 2$, as there is no optimal solution for a PLSE instance with a score of $n^2 - 1$ (cf. Theorem 6 in \cite{donovan2000completion}). 

\begin{algorithm}[h]
\small
\caption{Preprocessing procedure for graph reduction of the PLSE problem}
\begin{algorithmic}[1]\label{algo_preprocess}
\STATE \ 
\STATE \bf{Input}: \normalfont{A Latin square graph $G=(V,E)$ with some vertices already colored, each vertex $v$'s color domain $D(v)$.}
\STATE \bf{Output}: \normalfont{A reduced graph and the number $l$ of cells impossible to fill.}
\STATE
\FOR {each vertex $v \in V$ with singleton color domain $D(v) = \{k\}$}
\STATE $V \leftarrow V - \{v\}$ \hfill // Remove this colored vertex $v$ from the graph
\STATE $E \leftarrow E - \{\{u,v\} \in E \}$ \hfill // Remove the edges linked to $v$.
\FOR {each uncolored $u\in V$ adjacent to $v$}
\STATE $D(u) \leftarrow D(u)-\{k\}$ \hfill // Remove color $k$ from the color domain $D(u)$
\ENDFOR
\ENDFOR
\STATE
\STATE $l = 0$
\FOR {each $v \in V$}
\IF{$D(v) = \{0\}$}
\STATE $l = l + 1$
\STATE $V \leftarrow V - \{v\}$ \hfill // Remove this node impossible to color
\STATE $E \leftarrow E - \{\{u,v\} \in E \}$ \hfill // Remove the edges linked to $v$.
\ENDIF
\ENDFOR
\end{algorithmic}

\end{algorithm}

\begin{figure}[!t]
    \centering
    \includegraphics[width=0.65\textwidth]{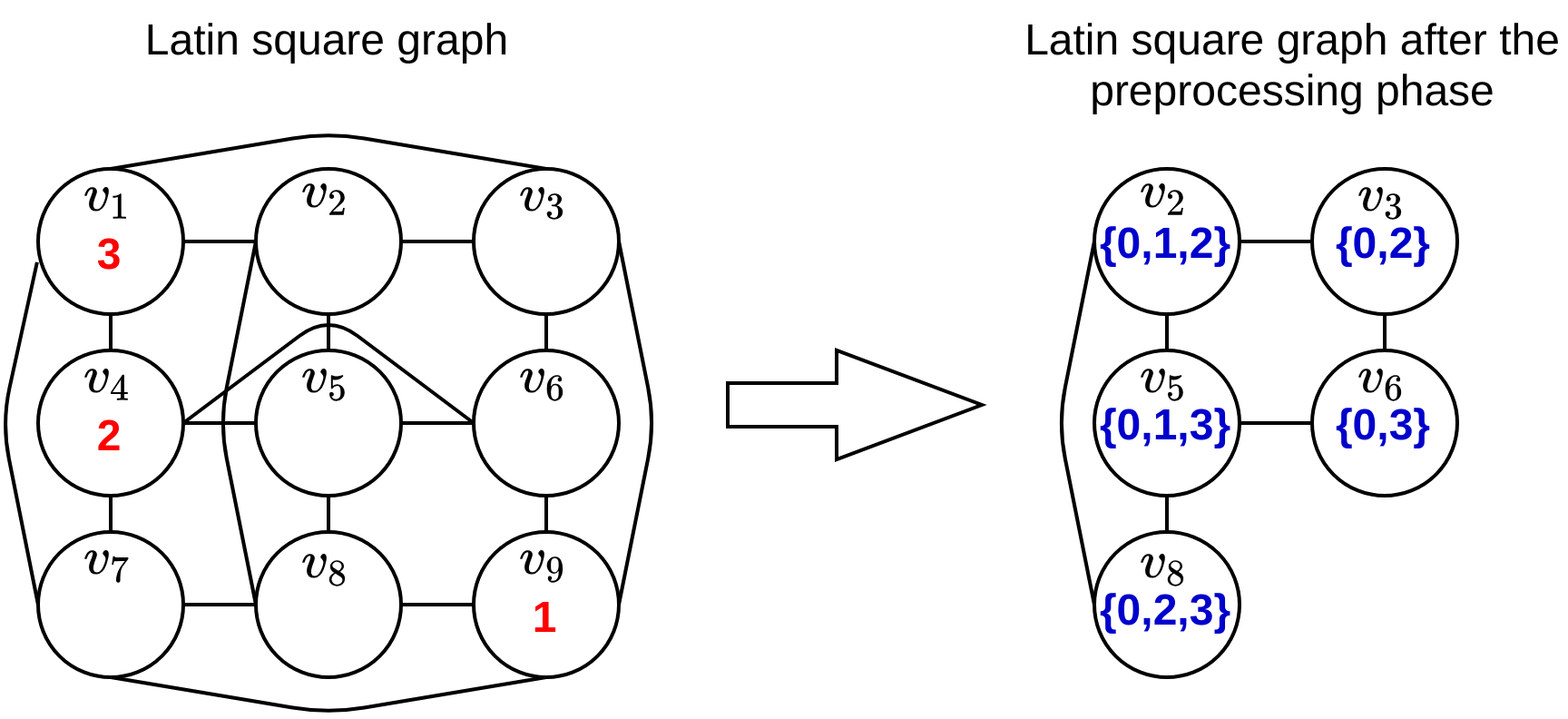}
    \caption{\label{fig:PLSE_3} Preprocessing of a Latin square graph with $n=3$. }
\end{figure}

Figure \ref{fig:PLSE_3} (Right) displays the reduced graph of the Latin square graph shown in Figure \ref{fig:PLSE_2}. Numbers in accolades indicate the color domain $D(v_i)$ of each vertex $v_i$. In addition to the three precolored vertices $v_1, v_4, v_9$, vertex $v_7$ is also removed because its color domain is $D(v_7)=\{0\}$. Therefore, $l = 1$, leading to an upper bound $3^2-2 = 7$. Since this upper bound is equal to the lower bound of the two solutions of Figure \ref{fig:PLSE_1}, these two solutions are proven to be optimal for the given PLSE instance (i.e., the maximum of 7 filled cells / colored vertices or the minimum of 2 unfilled cells / uncolored vertices). 

\subsection{Special features of the transformed coloring problem} \label{specificity}

One observes two special features of the graph coloring problem transformed from the PLSE. 

\begin{figure}[!t]
    \centering
    \includegraphics[width=0.65\textwidth]{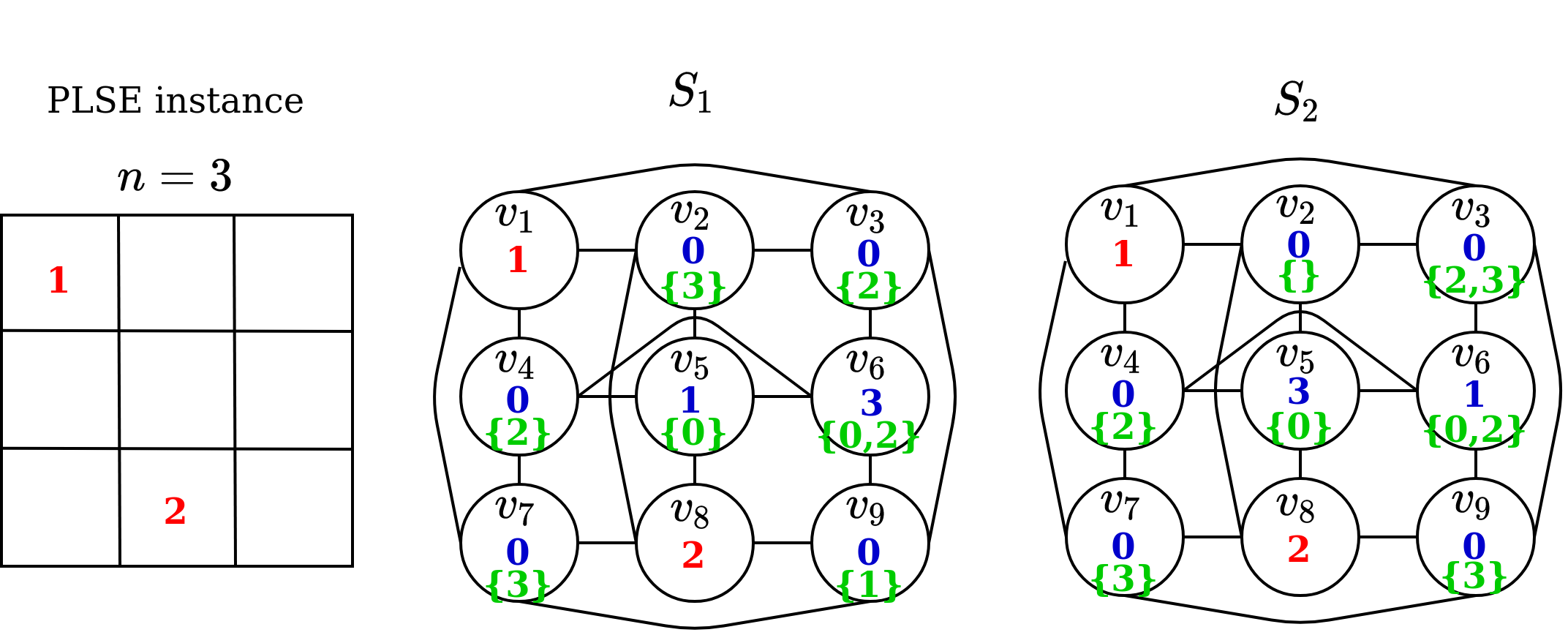}
    \caption{\label{fig:PLSE_4} Two legal solutions $S_1$ and $S_2$ of the PLSE instance. The two solutions are the same except that the colors 1 and 3 are swapped.}
\end{figure}

First, the Latin square graph coloring problem is a list-coloring problem  \cite{lewis2015guide}, where the permissible colors of a vertex are limited to a list of colors in \{0, 1, \dots, n\}, instead of the whole set \{0, 1,...,n\}.  Therefore, contrary to the standard graph coloring problem, candidate solutions are in general not invariant by permutation of colors. For example, in the legal partial coloring shown on Figure \ref{fig:PLSE_2} on the right, it is impossible to swapping colors 2 and 3 as the color 2 is not in the domain of the vertex $v_6$. Moreover, even a permissible color exchange  between two  colorings is not generally neutral. For example, consider the two legal solutions $S_1$ and $S_2$ displayed on Figure \ref{fig:PLSE_4}, where the pre-filled colors are  in red, assigned colors are in blue and possible color changes are in green. The solution $S_2$ is the same as the solution $S_1$ except that the colors 1 and 3 are swapped. After this swap, it becomes impossible to change the color of the vertex $v_2$ in $S_2$ while it was possible in $S_1$. $S_1$ and $S_2$ are thus two different candidate solutions for the PLSE, while they represent the same coloring for the conventional graph coloring problem. This observation implies that for this list-coloring problem, solutions are not invariant by permutation of the colors. As a result, the so-called set-theoretic partition distance \cite{porumbel2011efficient}, which is usually used to measure the distance between two solutions for graph coloring \cite{hao2012memetic,moalic2018variations}, is not meaningful for the list-coloring problem. Instead, the Hamming distance $D^H$ is more suitable to measure the distance between solutions for our coloring problem (cf. Section \ref{PopulationUpdate}).

Secondly, the partial list-coloring from the PLSE aims to find a legal coloring such that the objective function $f(S)$ defined by equation (\ref{eq:fitness}) (number of uncolored vertices) is minimized. Therefore, it is critical that the algorithm is able to decide which vertices are to be left uncolored when it is impossible to color all the vertices of the graph.

For these reasons, we introduce an algorithm specifically designed to solve the list-coloring problem of Latin square graphs of the PLSE. This algorithm, presented in the next section, can also be applied to solve the related Latin square completion problem (LSC).

\section{Massively parallel memetic algorithm}\label{sec:MPMA}

We describe in this section the massively parallel memetic algorithm (MPMA) for coloring Latin square graphs.

\subsection{Search space and evaluation function} \label{search_space}

The enlarged search space $\Omega$ explored by the MPMA algorithm is composed of the legal, illegal and potentially partial candidate solutions.  

Let $G=(V,E)$ be the reduced Latin square graph with $|V|$ vertices $\{v_1,\dots, v_{|V|}\}$, and color domains $D(v_i) \subseteq \{0,1,...,n\}$ ($i = 1,\dots,|V|$) obtained after the preprocessing phase. Then the space $\Omega$ is given by:
\begin{equation}
\Omega= \{S:V \rightarrow \{D(v_1),\dots,D(v_{|V|})\}\}
\end{equation}
The MPMA algorithm aims to find a legal, possibly partial solution $S$ (with $c(S)= 0$) of the Latin square graph with the minimum number of uncolored vertices $f(S)$  (for functions $f$ and $c$, see Section \ref{Latin square graph}).

Following the general idea of penalty function for constrained optimization \cite{chen2016hybrid,lai2018two,sun2018adaptive}, we define the following extended evaluation function $F$ (to be minimized) to assess the quality (fitness) of a candidate solution $S \in \Omega$:
\begin{equation}
F(S) = f(S) + \phi c(S)
\label{eq:extended_fitness}
\end{equation}
\noindent where $\phi > 0$ is a penalty parameter controlling the impact of the constraint function $c$ on the overall score. Generally, decreasing the value of $\phi$ favors solutions with less uncolored vertices and more conflicts, while increasing its value promotes legal (conflict-free) and partial colorings. If $\phi$ is set to the value of 1, $x$ uncolored vertices and $x$ conflicts contribute equally to the quality of the solution. 


\subsection{Main scheme}

The proposed MPMA algorithm relies on the population-based memetic framework \cite{MABook2012}, which has been applied to several graph coloring problems \cite{lu2010memetic,jin2014memetic,moalic2018variations,SunHao2020}. It's worth noting that these memetic algorithms typically use a small population with no more than 20 individuals and are elitist evolutionary algorithms. As such, each generation usually creates one or two offspring solutions via a crossover operator, which are further improved by a local search procedure.

The massively parallel memetic algorithm proposed in this work uses a very large population $P$ ($|P|\geq10^4$), whose individuals evolve in parallel in the search space. This approach secures a desirable high degree of population diversity, which in turn favors a large exploration of the candidate solutions. In order to take advantage of this large population, we leverage the computing power of modern GPUs to perform parallel computations at each generation: local searches, distance evaluations and crossovers. The only part which remains sequential is the population update operation that merges the current population and the offspring population to create the next population. 

\begin{algorithm}
\small
\caption{Massively parallel memetic algorithm for Latin sqaure graph coloring}\label{algo_memetic}
\begin{algorithmic}[1]
\STATE \bf{Input}: \normalfont{Reduced Latin square graph $G = (V, E)$, population size $p$, color domain $D(v)$
of each vertex $v \in V$.}
\STATE \bf{Output}: \normalfont{The best legal  partial coloring $S^*$ found}
\STATE \normalfont{$P = \{S_1,\dots,S_p\}$ $\leftarrow$ population\_initialization} 
\STATE $S^*= \emptyset$ and $e(S^*)= |V|$.
\STATE $\{S^O_1,\dots,S^O_p\} \leftarrow \{S_1,\dots,S_p\}$
\REPEAT
\FOR {$i = \{1, \dots, p \}$, \textbf{in parallel}}
\STATE $S'_i \leftarrow \text{two-phase\_local\_search}(S^O_i)$ \hfill $/*$ Section \ref{PartialTabuCol}
\ENDFOR
\STATE $S'^* = \text{argmin} \{f(S'_i), i= 1, \dots, p \}$
\IF{$f(S'^*) < f(S^*)$}
\STATE $S^* \leftarrow S'^*$
\ENDIF
\STATE $D \leftarrow \text{distance\_computation} (S_1, \dots, S_p,S'_1, \dots, S'_p )$ \hfill $/*$ Section \ref{PopulationUpdate}
\STATE $\{S_1, \dots, S_p \} \leftarrow \text{pop\_update} (S_1, \dots, S_p,S'_1, \dots, S'_p , D)$ \hfill $/*$ Section \ref{PopulationUpdate}
\STATE $\{S^O_1, \dots, S^O_p \} \leftarrow \text{build\_offspring}(S_1, \dots, S_p, D)$ \hfill $/*$ Section \ref{Crossovers}
\UNTIL{stopping condition met}
\RETURN $S^*$
\end{algorithmic}
\end{algorithm}

\begin{figure}[!h]
    \centering
    \includegraphics[width=0.65\textwidth]{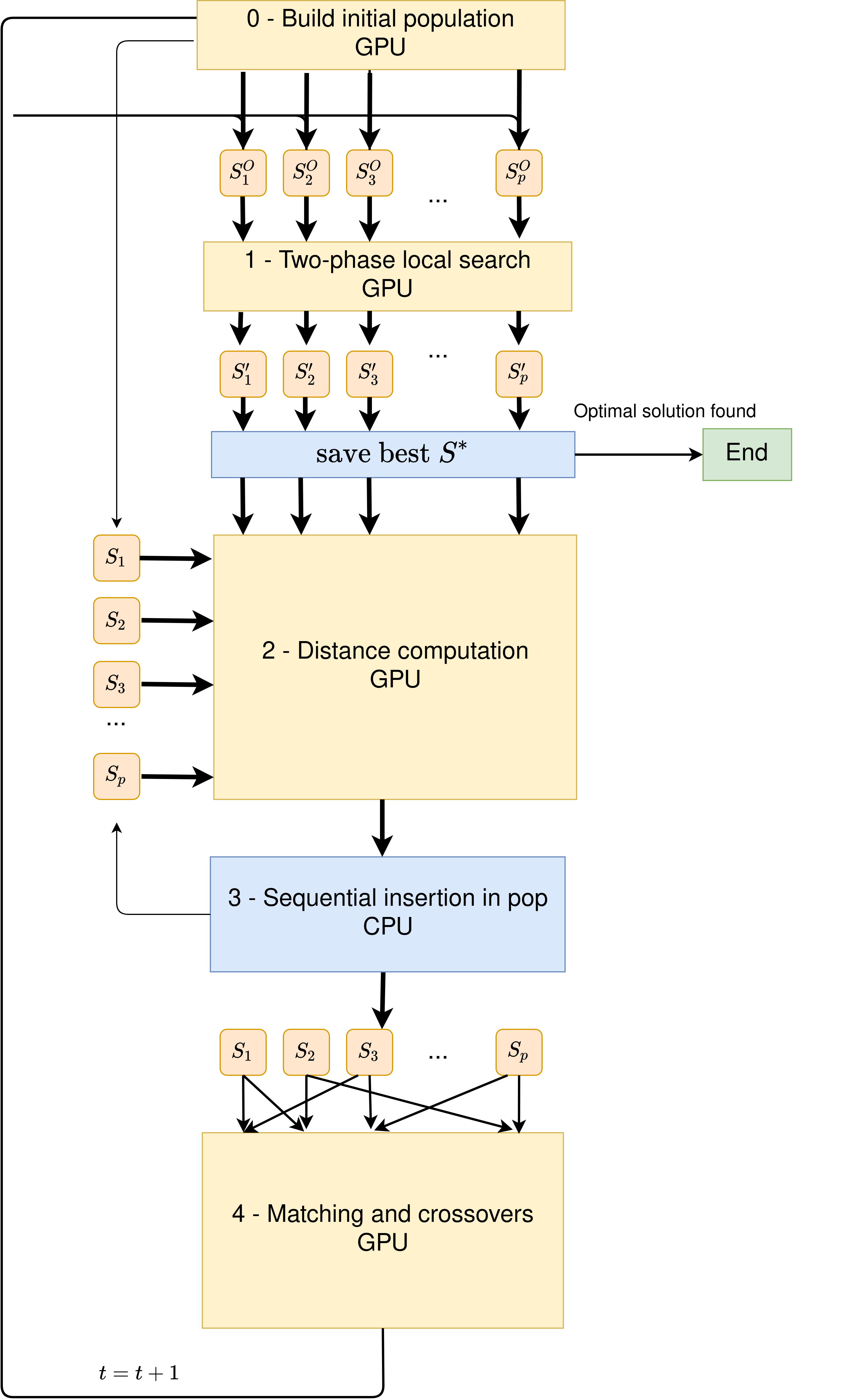}
    \caption{\label{fig:GPU_memetic} General scheme of the MPMA algorithm.}
\end{figure}

The  algorithm  takes  a  reduced  Latin  square  graph $G$ (cf. Section \ref{preprocess_Latin_square}) as  input  and  tries  to  find  a  legal, possibly  partial,  coloring  with a minimum number of uncolored vertices. The pseudo-code of MPMA is shown in Algorithm \ref{algo_memetic}, while its flowchart is displayed in Figure \ref{fig:GPU_memetic}. At the beginning, all the individuals of the population are initialized at random in parallel. Then, the algorithm repeats a loop (generation) until a stopping criterion (e.g., a cutoff time limit or a maximum of generations) is met. Each generation $t$ involves the execution of four components:

\begin{enumerate}
\item The $p$ individuals (illegal $n$-colorings) of the current population are simultaneously improved by running  in parallel a two-phase local search (cf. Section \ref{PartialTabuCol}) to minimize the fitness function $f$ (uncolored vertices) and the constraint function $c$ (conflicting vertices).
\item The distances between all pairs of the existing individuals and the individuals improved by local search are computed in parallel (cf. Section \ref{PopulationUpdate}). 
\item Then the population updating procedure (cf. Section \ref{PopulationUpdate}) merges the $2p$ existing and new individuals to update the population, by taking into account the fitness $f$ of each individual (number of uncolored vertices) and the distances between individuals in order to maintain a healthy diversity of the population.
\item Finally each individual is matched with its nearest neighbor individual in the population and $p$ crossovers are performed in parallel to generate $p$ offspring solutions (cf. Section \ref{Crossovers}), which are improved by the parallel iterated local search during the next generation ($t+1$).
\end{enumerate}

The algorithm stops when a predefined time condition is reached or an optimal solution $S^*$ is found. $S^*$ is an optimal solution if 1) $c(S^*)= 0$, $f(S^*)=0$, and $l \neq 1$ (i.e., all empty cells are filled), or 2) $c(S^*)= 0$, $f(S^*)=1$, and $l = 1$ (the tightest upper bound is reached, see Section \ref{preprocess_Latin_square}). If the algorithm does not find an optimal solution when it stops, it returns the best legal solution $S^*$ (with $c(S^*)= 0$) found so far, with a number of unfilled cells $f(S^*) > 0$. Then the score $n^2 - l - f(S^*)$ is a lower bound of the given PLSE instance.

\subsection{Parallel two-phase local search} \label{PartialTabuCol}

MPMA employs a two-phased partial legal and illegal tabu search (PLITS) to simultaneously improve in parallel the individuals of the current population. Specifically, PLITS relies on the tabu search metaheuristic to explore candidate solutions of the space $\Omega$ guided by the extended fitness function $F$ given by equation (\ref{eq:extended_fitness}). Indeed, tabu search is a popular method for graph coloring \cite{hertz1987using,sun2018adaptive,wang2018tabu} and often used as the local optimization components of memetic algorithms \cite{jin2019solving,moalic2018variations,galinier1999hybrid}.

Given a solution $S=\{V_0,V_1,V_2,...,V_n\}$, PLITS uses the one-move operator to displace a vertex $v$ from its current color class $V_i$ to a different color class $V_j$ such that $i\neq j$ and $j \in D(v)$, leading to a neighboring solution denoted as $S \oplus <v, V_i, V_j >$. Let $\mathcal{C}(S)$ be the set of conflicting vertices in $S$, i.e., $\mathcal{C}(S) = \{ v \in V_i : 1 \leq i \leq n, \exists u \in V_i, (u,v) \in E, u \neq v \}$. To make the examination of candidate solutions more focused, PLITS only considers the uncolored vertices in $V_0$ and conflicting vertices in $\mathcal{C}(S)$ for color changes. 

The one-move neighborhood applied to the uncolored vertices of $S$ is given by:
\begin{equation*}
    N_0(S) = \{S \oplus <v, V_0, V_j > : v \in V_0, 1 \leq j \leq n, j \in D(v) \}
\end{equation*}
The one-move neighborhood applied to the conflicting vertices of $S$ is given by:
\begin{align*}
N_c(S) = &  \{S \oplus <v, V_i, V_j > : v \in \mathcal{C}(S), v \in V_i, 1 \leq i \leq n, \\ &  0 \leq j \leq n, j \in D(v), i \neq j  \}
\end{align*}
Notice that a conflicting (colored) vertex can be moved to the set $V_0$ by the one-move operator, becoming thus uncolored.

PLITS explores the global one-move neighborhood:
\begin{equation}
N(S) =  N_0(S) \cup N_c(S)
\label{eq:neighborhood}
\end{equation}
Specifically, PLITS makes transitions between various partial $n$-colorings with the help of the neighborhood $N(S)$ and the extended evaluation function $F$. It iteratively replaces  the  current  solution $S$ by  a neighboring solution $S'$ taken from $N(S)$, until a stopping condition is met.  At each iteration, a best admissible neighboring solution $S'$ is selected to replace $S$. After each iteration, the corresponding one-move is recorded in the tabu list to prevent the search from returning to a previously visited solution for the next $T$ iterations (tabu tenure). Like \cite{galinier1999hybrid}, the tabu tenure depends on the number of vertices eligible for the one-move operator (i.e., $|V_0| + |\mathcal{C}(S)|$ in our case) and is set to the value of $L + \alpha (|V_0| + |\mathcal{C}(S)|)$, where $L$ is a random integer from $[0;9]$ and $\alpha$ is a parameter set to 0.6.
 
A neighboring solution $S'$ is considered to be admissible if it is not forbidden by the tabu list or if it is better (according to the extended evaluation function $F$) than the best solution found so far. The neighborhood evaluations are performed incrementally using the streamline technique of \cite{galinier1999hybrid}. As shown in Algorithm \ref{algo_PLITS}, we run in parallel the PLITS procedure on the GPU to raise the quality of the current population. 


\begin{algorithm}[h]
\small
\caption{Parallel partial legal and illegal tabu search}\label{algo_PLITS}
\begin{algorithmic}[1]
\STATE \bf{Input}: \normalfont{Population  $P= \{S_1, \dots, S_p \}$, depth of tabu search $nbIter_{TS}$, color domain $D(v)$ of each vertex $v \in V$. }
\STATE \bf{Output}: \normalfont{Improved population $P'= \{S'^*_1, \dots, S'^*_p \}$}.
\FOR {$i = \{1, \dots, p \}$, \textbf{in parallel}}
\STATE $S'^*_i \leftarrow S_i$ \hfill $/*$ Records the best solution found so far on each local thread.
\ENDFOR
\STATE $iter = 0$
\FOR {$i = \{1, \dots, p \}$, \textbf{in parallel}}
\FOR {$t = \{1, \dots, nbIter_{TS} \}$}
\STATE Choose a neighboring solution $S'_i \in N(S_i)$ which is not forbidden by the tabu list or better than $S_i$ (according to the extended evaluation function $F$).
\STATE $S_i \leftarrow S'_i $
\IF{$F(S'_i) < F(S'^*_i)$}
\STATE $S'^*_i \leftarrow S'_i$
\ENDIF
\ENDFOR
\ENDFOR
\STATE \textbf{return} $P'= \{S'^*_1, \dots, S'^*_p \}$
\end{algorithmic}
\end{algorithm}

The PLITS procedure is performed in two phases with different search focuses. The first phase favors a large exploration of candidate solutions by setting $\phi$ to the value of 0.5 and performs $nbIter_{TS} = 100*|V|$ iterations. The second phase focuses on resolving the conflicts in the solutions of the population to obtain $P$ legal colorings (with $c(S)=0$). For this purpose, $\phi$ is set to a large value of $|V|$ during $nbIter_{TS} = 2*|V|$ iterations.

After the local search, the best coloring $S'^*_i$ among the $p$ conflict-free colorings in terms of the objective function $f$ is used to update the recorded best solution $S^{*}$ if $S'^*_i < S^{*}$.

\subsection{Population update} \label{PopulationUpdate}

The $p$ new legal colorings from the PLITS procedure are used to update the population. For this, MPMA maintains a $p \times p$ matrix to record all the distances between any two solutions of the population. This symmetric matrix is initialized with the $p \times (p-1)/2$ pairwise distances computed  for each pair of individuals in the initial population, and then updated each time a new individual is inserted in the population.

To merge the $p$ new solutions and the $p$ existing solutions, MPMA needs to evaluate (i) $p \times p$  distances between each individual in the population $P=\{S_1,\dots,S_p\}$ and each improved offspring individual in $P' = \{S'_1,\dots,S'_p\}$ and (ii) $p \times (p-1)/2$ distances between all the pairs of individuals in $P'$. All the $p\times p + p\times(p-1)/2$ distance computations are independent from one another, and are performed in parallel on the GPU (one computation per thread).

Given two colorings $S_i$ and $S_j$, MPMA uses the Hamming distance $D(S_i,S_j)$ to measure the dissimilarity between $S_i$ and $S_j$, which corresponds to the number of vertices that are colored differently in $S_i$ and $S_j$:
\begin{equation}
D(S_i,S_j) = |\{v \in V, S_i(v) \neq S_j(v) \}|
\end{equation}
Following \cite{hao2012memetic}, MPMA's population update procedure aims to keep the best  individuals, but also to ensure a minimum spacing between the individuals. The population update procedure (Algorithm \ref{pop_update}) greedily adds one by one the best individuals of $P^{all} =\{S_1, \dots, S_p \} \cup  \{S'_1, \dots, S'_p\}$ in the population of the next generation $P_{t+1}$ until $P_{t+1}$ reaches $p$ individuals, such that $D(S_i,S_j) > |V|/\gamma$ ($\gamma >1.0$ is a parameter), for any $S_i,S_j \in P_{t+1}$, $i \neq j$. 

\begin{algorithm}[!t]
\small
\caption{Sequential population update procedure}\label{pop_update}
\begin{algorithmic}[1]
\STATE \bf{Input}: \normalfont{Population  $P_t= \{S_1, \dots, S_p \}$ (generation $t$) and  offspring population  $P'=\{S'_1, \dots, S'_p\}$} (generation $t$)
\STATE \bf{Output}: \normalfont{Updated population $P_{t+1}$  (generation $t+1$)} 
\STATE $P_{t+1} = \emptyset$ \hfill $/*$ Initilize new population
\STATE $P^{all} = P_t \cup P'$ \hfill $/*$ Merge existing and improved new solutions
\STATE $S^{best} = \text{argmin}_{S \in P^{all}}\ e(S)$ \hfill $/*$ Identify the best legal solution in $P^{all}$
\STATE $P_{t+1} = P_{t+1} \cup \{S^{best}\}$ \hfill $/*$ Add $S^{best}$ in $P_{t+1}$
\STATE $P^{all} = P^{all} \setminus \{S^{best}\}$  \hfill $/*$ Remove $S^{best}$ from $P^{all}$
\STATE $/*$ Add $n$-colorings in $P_{t+1}$ until it contains the $p$ best solutions of $P^{all}$ with the condition that $D(S_i,S_j) > |V|/10$, for all $S_i,S_j \in P_{t+1}$, $i \neq j$  
\WHILE{$|P_{t+1}| < p$}
\STATE $S^{best} = \text{argmin}_{S \in P^{all}}\ e(S)$
\STATE $dist =  \text{min}_{A \in P_{t+1}}\ D(S^{best},A)$
\IF{$dist > |V|/10$}
\STATE $P_{t+1} = P_{t+1} \cup \{S^{best}\}$
\STATE $P^{all} = P^{all} \setminus \{S^{best}\}$
\ENDIF
\ENDWHILE
\RETURN $P_{t+1}$
\end{algorithmic}
\end{algorithm}

\subsection{Parent matching and crossover} \label{Crossovers}

At each generation, the MPMA algorithm performs in parallel $p$ crossovers to generate $p$ offspring solutions. For this, MPMA uses each existing solution in the current population as the first parent and selects another existing solution as the second parent with a specific parent matching strategy. The idea is to ensure that each individual in the population has a chance to transmit some \textit{genetic information} to the next generation while encouraging the creation of diversified offspring.

\subsubsection{Parent matching strategy}  \label{sec:matching}

The population update strategy presented in the last section ensures that the individuals of the next population are of high quality, but also sufficiently distanced. This property provides a first basis to guarantee that for each of the $p$ crossovers, we can find a second parent that is sufficiently distanced from the first parent. This contributes to building diversified offspring solutions that are different from their parents and thus helps the algorithm to continually explore new areas in the search space.

However, as we use a very large population, individuals can be highly different and share very little information. Indeed, we experimentally observed that the average pairwise distance in the population is usually very large, around $0.7 \times |V|$ even after many generations. Meanwhile, a study in \cite{moalic2018variations} showed that for the standard graph coloring problem, crossing-over two highly different parents results in offspring of poor quality because no meaningful shared information can be transmitted from parents to offspring.

Thus, for each individual $S_i$ (i.e., the first parent), we choose, among the other individuals in the population, the nearest neighbor $S_j$ in the sense of the precomputed Hamming distance $D$, as the second parent.

\subsubsection{Parameterized asymmetric uniform crossover \label{sec:AUXcross}}

The popular greedy partition crossover (GPX) \cite{galinier1999hybrid} and its variants have proven to be very successful for several graph coloring problems including the conventional graph coloring \cite{lu2010memetic,moalic2018variations,malaguti2008metaheuristic} and sum coloring \cite{jin2014memetic}. GPX was also adapted to the related LSC, leading to maximum approximate group based crossover (MAGX) \cite{jin2019solving}. However, the GPX crossover has some limitations for the PLSE due to the fact that solutions are not invariant by permutations of color groups (cf. Section \ref{specificity}) and high-quality solutions do not share significant backbones (they are far away from each other, see Section \ref{sec:key_components}).
 
For the PLSE, we introduce a parameterized asymmetric uniform crossover (AUX), which is easy to compute for a very large population of individuals and allows the transmission of favorable parental features to the next generation.
 
Given a first parent $S_i$ and a second parent $S_j$, an offspring solution $S^O_i$ is built such that each vertex $v$ receives the color of $S_i$ with probability $p_{ij}$ and the color of $S_j$ with probability $1 - p_{ij}$. The probability $p_{ij}$ depends proportionally on the Hamming distance between the parents $S_i$ and $S_j$ and is given by:
 \begin{equation}
    p_{ij} = 1 - \frac{|V|}{\beta \times D(S_i,S_j)}
    \label{eq:probaPij}
 \end{equation}
 \noindent where $\beta > 1.0$ is a real parameter controlling the degree of diversity of the resulting offspring.
 
 As $|V|/\gamma$ is the minimum spacing between two individuals in the population (cf. Section \ref{PopulationUpdate}), we set $\beta$ such that $\beta > \gamma >0$, in order to have $\forall i,j \in \llbracket 1, \dots, p \rrbracket^2, i \neq j,\ |V|/\beta < D(S_i,S_j)$. This ensures that $\forall i,j \in \llbracket 1, \dots, p \rrbracket^2, \ 0 < p_{ij} < 1$.
 
Notice that when $p_{ij}$ is fixed to the value of 0.5, we obtain the classical Uniform Crossover (UX) \cite{Syswerda89}. With the UX crossover, the resulting offspring is on average equidistant from both parents. However, as we empirically show in Section \ref{sec:experiments}, the UX crossover does not work well for the PLSE (it is too much disruptive). The proposed AUX crossover uses the probability $p_{ij}$ to make itself more conservative by considering the distance between two parents. Specifically, if two parents are similar (with a small distance), the offspring can equally inherit information from the parents. On the contrary, if the parents are very different (with a large distance), it is preferable to conserve more information from one parent (the first parent) to avoid an offspring solution that is far away from both parents. AUX achieves this goal by adjusting the coefficient $\beta$ which influences the probability.

\begin{algorithm}[!t]
\small
\caption{Parallel asymmetric uniform crossover AUX}
\begin{algorithmic}[1]\label{crossoverAUX}
\STATE \bf{Input}: \normalfont{Population  $P= \{S_1, \dots, S_p \}$, with $S_i = (V^i_0, V^i_1,\dots, V^i_n)$, for $i = 1,\dots,p$. }
\STATE \bf{Output}: \normalfont{Offspring population $P^O= \{S^O_1, \dots, S^O_p \}$} 
\FOR {$i = 1, \dots, p$, in parallel}
\STATE $S_j \leftarrow $ Find and make a copy of the nearest neighbor of $S_i$ from $P$ according to the distance $D$ such that $i \neq j$ and such that this crossover $(i,j)$ has not been tested yet.
\STATE $p_{ij} = 1 - \frac{|V|}{\beta \times D(S_i,S_j)}$
\FOR {$l = \{1, \dots, |V| \}$}
\STATE With probability $p_{ij}$, $S^O_i(v_l) = S_i(v_l)$
\STATE Otherwise $S^O_i(v_l) = S_j(v_l)$
\ENDFOR
\ENDFOR
\RETURN $P^O$
\end{algorithmic}
\end{algorithm}

For two given parents $S_i$ and $S_j$, the expected distance between the offspring $S^O_i$ and its first parent $S_i$  is   $\bar{D}(S_i,S^O_i) = |V|/\beta $.  The expected distance between the offspring $S^O_i$ and its second parent $S_j$ is   $\bar{D}(S_j,S^O_i) = D(S_i,S_j) - |V|/\beta$. If we choose $\beta \geq 2 \gamma$, $\bar{D}(S_i,S^O_i) \geq \bar{D}(S_j,S^O_i)$ always holds. As such, in average the child preserves more genetic information from the first parent compared to the second parent. Given that MPMA uses every individual in the current population as the first parent, all individuals are offered the same chance to transmit a large part of their genetic information to their offspring, leading to a large coverage of the search space. 

\begin{figure}[!h]
    \centering
    \includegraphics[width=0.65\textwidth]{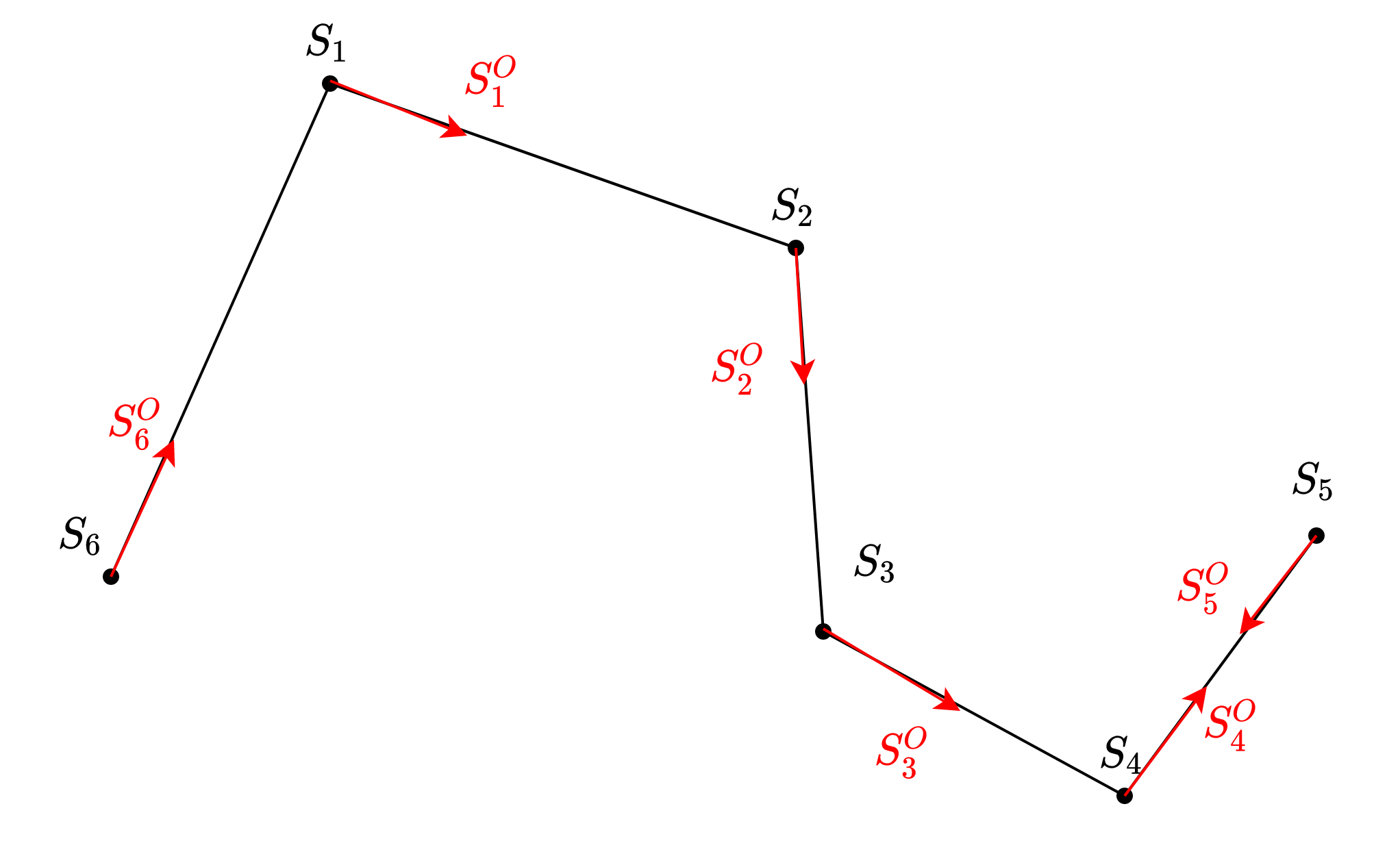}
    \caption{\label{fig:GPU_memetic} Resulting offspring individuals  $\{S^O_i\}_{i=1}^6$ (in red) generated from the population $\{S_i\}_{i=1}^6$ (in black). } 
    \label{fig:offspring_pop}
\end{figure}

Figure \ref{fig:offspring_pop} illustrates the creation of six offspring solutions $\{S^O_i\}_{i=1}^6$ (in red) generated from the population $\{S_i\}_{i=1}^6$ (in black). In this case, the offspring $S^O_1$ to $S^O_6$ are respectively generated from the ordered pairs of parents $(S_1,S_2)$, $(S_2,S_3)$, $(S_3,S_4)$, $(S_4,S_5)$, $(S_5,S_4)$, $(S_6,S_1)$.

As one notices, each offspring is situated in between its two parents in the search space and always closer to its first parent (in terms of the Hamming distance). The norm of each translation vector is equal to $|V|/\beta$ in average.

The overall parent matching and the AUX crossover are summarized in Algorithm \ref{crossoverAUX}. All the $p$ crossover operations are performed in parallel on individual GPU threads.

\subsection{MPMA implementation on graphic processing units} \label{sec:GPUimplem}

MPMA was programmed in Python with the Numba library for CUDA kernel implementation. It is specifically designed to run on GPUs.  In this work we used a V100 Nvidia graphic card with 32 GB memory.

\begin{figure}[h]
    \centering
    \includegraphics[width=0.65\textwidth]{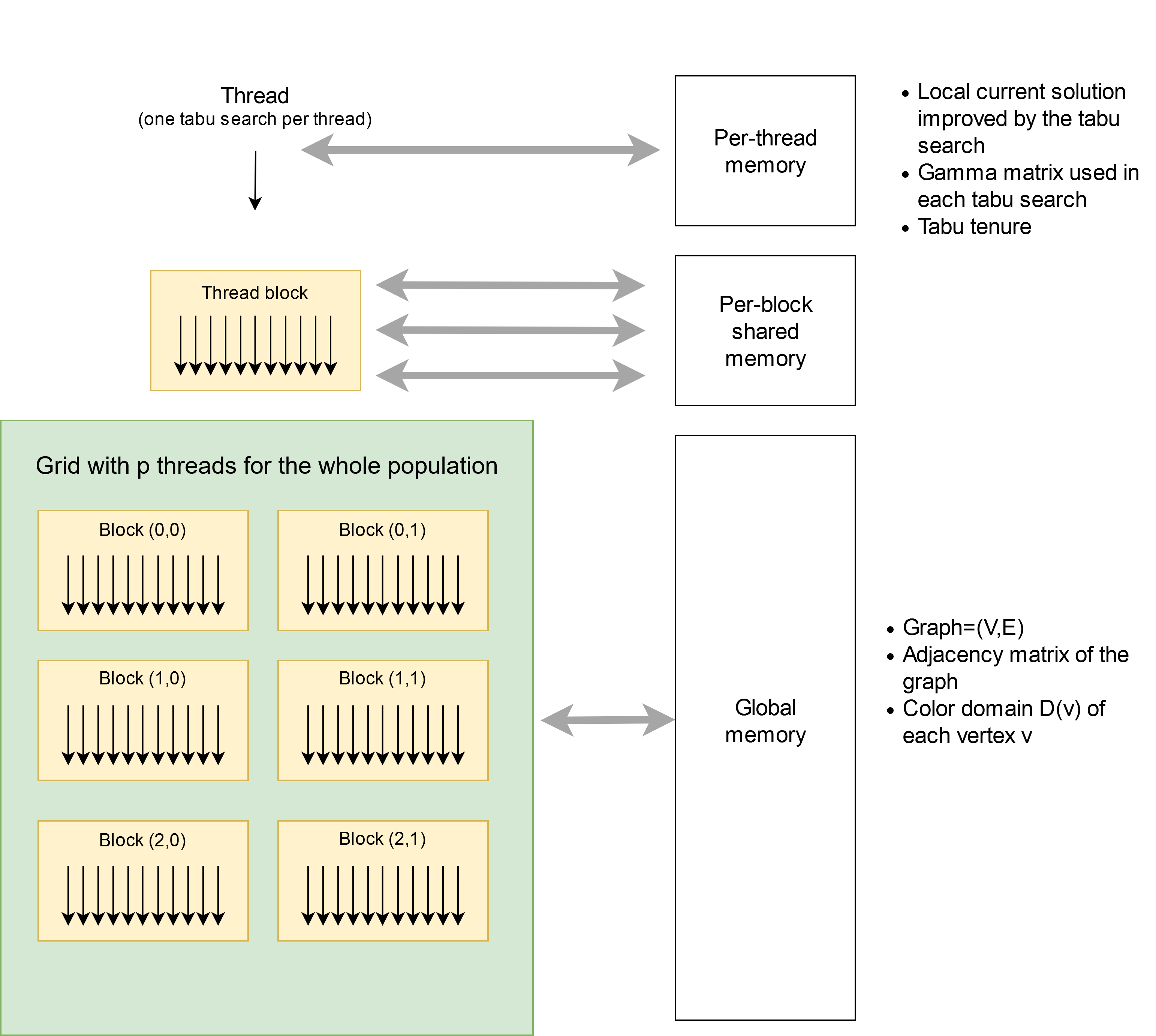}
    \caption{\label{fig:gpu_memory} Parallel tabu searches launched on GPU grid.}
\end{figure}

Figure \ref{fig:gpu_memory} shows the organization of the threads on the GPU grid and the memory hierarchy on the GPUs used to run the $p$ tabu searches in parallel for the whole population at each generation.  Each of the $p$ tabu searches (cf. Section \ref{PartialTabuCol}) is run on a single thread. For fast memory access, a per-thread local memory is used  to store specific local information such as the current solution being ameliorated and the tabu tenure. The threads are grouped by block of size 64 and launched on the GPU grid. No per-block shared memory is used because all the local searches performed in each block are independent from each other.  However, a global memory is employed to store general information about the graph such as its adjacency matrix of the graph and the color domain of each vertex to avoid information duplication. All these $p$ tabu searches are launched with a CUDA kernel function and the best results obtained during each tabu search are transferred to the CPU after synchronization.

The same type of kernel function on the GPUs is used to compute in parallel the $p\times p + p\times(p-1)/2 $ distance calculations (cf. Section \ref{PopulationUpdate}) and the $p$ crossovers (cf. Section \ref{Crossovers}) at each generation. However, some operations such as the best solution saving procedure and the population update procedure (cf. Section \ref{PopulationUpdate}) are performed on the CPU as they cannot be parallelized.

\subsection{A variant of the algorithm for highly constrained instances}\label{partialMPMA}

As shown in Section \ref{sec:experiments}, the MPMA algorithm excels on under-constrained to moderately over-constrained PLSE instances with a filled ratio $r$ below $80\%$. However, its performance slightly deteriorates on highly constrained instances when $r \geq 80 \%$. For these cases, we observed that better results can be reached by directly minimizing the number of uncolored vertices (i.e., fitness $f(.)$ of Section \ref{Latin square graph}) in the space of legal (i.e., conflict-free) partial colorings. For these highly constrained instances, we create a simplified MPMA variant called Partial-MPMA that works with legal partial colorings (instead of conflicting colorings) and makes the following two changes in MPMA.

\begin{itemize}
    \item A greedy conflict removal procedure is applied to repair each offspring solution into a legal partial coloring. For this, the vertex which is conflicting with the largest number of vertices is uncolored first (i.e., reassigned the color $0$), followed by the vertex with the second largest conflicts and so on. This process continues until a partial conflict-free coloring is reached.
    \item The two-phase tabu search procedure of Section \ref{PartialTabuCol} is replaced by the PartialCol coloring algorithm of \cite{blochliger2008graph} adapted to the list-coloring problem. This PartialCol algorithm uses tabu search to explore the space of legal partial colorings by minimizing the number of uncolored vertices. 
\end{itemize}


\section{Experimental results \label{sec:experiments}}

This section is dedicated to a computational assessment of the MPMA algorithm for solving the partial Latin square extension problem, by making comparisons with the state-of-the-art methods. Additional results are presented in Appendix \ref{sec:results_LSC} for the related Latin square completion problem.

\subsection{Benchmark instances} \label{Benchmark} 

We carried out extensive experiments on the 1800 PLSE benchmark instances introduced in \cite{haraguchi2016iterated}. These instances are parametrized by the grid order $n \in \{50,60,70\}$ and the ratio $r \in \{0.3,0.4,\dots,0.8\}$ of pre-filled cells in the $n \times n$ grid. Given $(n,r)$ and starting from an empty $n \times n$ grid, a PLSE instance was constructed by repeatedly assigning a different symbol in an empty cell chosen randomly so that the Latin square condition is respected and until $r \times n^2$ cells are assigned symbols. For each $(n,r)$ combination, 100 instances are available. Note that such a PLSE instance does not always admit a complete solution (i.e., some cells must be left unfilled). This is typically the case for relatively strongly constrained instances when $r > 60$ (i.e., when at least 60\% cells are pre-filled). Moreover, as shown in \cite{gomes2002completing,haraguchi2016iterated}, under-constrained instances ($r\le 0.5$) and over-constrained instances ($r>0.7$) are easier than medium-constrained instances with $r$ between $0.6$ and $0.7$.

It is clear that $n^2$ is an upper bound for these instances (all cells are filled). When the grid cannot be fully filled, a safe upper bound is given in \cite{donovan2000completion}, corresponding to $n^2 - 2$ (all but 2 cells are filled). This bound indicates that if a grid cannot be completed, at least two cells will be left unfilled.

Like \cite{jin2019solving}, we first convert these instances to Latin square graphs and apply the preprocessing algorithm of Section \ref{preprocess_Latin_square} to reduce them, leading to graphs with less than 500 vertices for $(n,r)=(50,0.8)$ and up to 3430 vertices for $(n,r)=(70,0.3)$. The preprocessing takes no more than several seconds.

\subsection{Parameter setting \label{sec:Parameter}} 

The population size $p$ of MPMA is set to $p=12288$, which is chosen as a multiple of the number of 64 threads per block. This large population size offers a good performance ratio on the Nvidia V100 graphics card that we used in our experiments, while remaining reasonable for pairwise distance calculations in the population, as well as the memory occupation on the GPU, especially when solving very large instances. A sensitivity experiment of the results with respect to the population size is presented in Section \ref{sec:key_components}. In addition to the population size, the parameter $\alpha$ of the tabu search is set to its classical value of 0.6 and the number of tabu iterations $nbIter_{TS}$ depends on the size $|V|$ of the graph. The parameter $\gamma$ for the minimum spacing between two individuals is set to $10$. The parameter $\beta$ for adjusting the distance of the offspring from their parents is fixed at 20.

Table \ref{table:parameters_MPMA} summarizes the parameter setting, which can be considered as the default and is used for all our experiments.

\begin{table}[!h]
\centering
\caption{Parameter setting in MPMA}
\scriptsize
\begin{tabular}{lll} 
Parameter & Description & Value\\
\hline
$p$ & Population size & 12288\\
$nbIter_{TS}$ & Number of iterations tabu search & $100 \times |V|$\\ 
$\alpha$ & Tabu tenure parameter  & 0.6\\
$\gamma$ & Parameter for the spacing between two individuals & 10 \\
$\beta$ & Parameter for the generation of offspring  & 20\\
\hline
\end{tabular}
\label{table:parameters_MPMA}
\end{table}

\subsection{Comparative results on the set of 1800 PLSE instances \label{sec:benchmarks}}

This section shows a comparative analysis on the 1800 PLSE instances with respect to the state-of-the-art methods.  Given the stochastic nature of the MPMA algorithm, each instance is independently solved 5 times. 

Table \ref{all_results_litterature} summarizes the computational results of MPMA compared to the best results in the literature reported in \cite{haraguchi2016iterated,jin2019solving}. For each instance MPMA was launched with a maximum of 100 billions of tabu search iterations. The reference methods include the 7 PLSE approaches in \cite{haraguchi2016iterated}: CPX-IP, CPX-CP, LSSOL, 1-ILS*, 2-ILS, 3-ILS and Tr-ILS*, where CPX-IP and CPX-CP are exact Integer Programming and Constraint Programming solvers from IBM/ILOG CPLEX, LSSOL denotes the tool LocalSolver. 1-ILS*, 2-ILS, 3-ILS and Tr-ILS* are four iterated local search algorithms with three different neighborhoods. We also cite the results of the recent MMCOL algorithm \cite{jin2019solving}, which is designed for the related LSC problem and reported results on the 1800 PLSE instances with an adapted version of MMCOL.

\begin{landscape}
\begin{table}

\centering
\scriptsize
\caption{Comparative results of MPMA and its Partial-MPMA variant with the state-of-the-art methods (CPX-IP, CPX-CP, LSSOL, 1-ILS*, 2-ILS, 3-ILS, Tr-ILS* in \cite{haraguchi2016iterated} and MMCOL in \cite{jin2019solving}) in terms of the average number of filled cells for each type of 100 PLSE instances of size $n \in \{50,60,70\}$ and ratio of pre-assigned symbols $r \in \{0.3,0.4,0.5,0.6,0.7,0.8\}$. Dominating  results are indicated in bold. \label{all_results_litterature}}

\begin{tabular}{ll|l|l|l|l|l|l|l|l|l|l} 

   \hline
 \multicolumn{2}{c|}{Instance}  &    \multicolumn{1}{|c|}{CPX-IP} & CPX-CP & LSSOL & 1-ILS* & 2-ILS & 3-ILS & Tr-ILS*  & MMCOL  &  \multicolumn{1}{|c}{MPMA} &  \multicolumn{1}{|c}{Partial-MPMA}   \\
   \hline
    $n$ & $r$  & $f_{best}$ & $f_{best}$ & $f_{best}$ & $f_{best}$ & $f_{best}$  & $f_{best}$ & $f_{best}$ & $f_{best}$  &  $f_{best}$  &  $f_{best}$ \\
    \hline
  \multirow{7}{*}{50} & 0.3  & 2496.03 & 2499.87 & 2496.35  & \textbf{2500*} & 2499.98 & 2499.96 & \textbf{2500*} & \textbf{2500*}  & \textbf{2500*}  & \textbf{2500*} \\
  & 0.4  & 2493.78 & 2498.02 & 2494.65 & 2499.98 & \textbf{2500*} & 2499.86 & \textbf{2500*} & \textbf{2500*} & \textbf{2500*}  & 2493.35\\  
  & 0.5  & 2488.52 & 2489.92  & 2492.96 & 2499.89 & 2499.95 & 2499.25 & \textbf{2500*} & \textbf{2500*} & \textbf{2500*} & 2491.82 \\ 
  & 0.6  & 2476.21 & 2478.87 & 2489.21 & 2496.23 & 2496.3 & 2494.67 & 2497.18  & 2499.64  & \textbf{2499.7} & 2485.64\\ 
  & 0.7 & 2446.4 & 2451.04 & 2463.45 & 2469.47 & 2469.78 & 2467.77 & 2470.07  & 2478.94  & \textbf{2484.38}  & 2466.95 \\ 
  & 0.8 & \textbf{2394.58} & 2388.1 & 2393.67 & 2394.14 & 2394.11 & 2394.09 & 2394.14  & 2364.61  & 2393.24 &  \textbf{2394.58} \\
 \hline
  \multirow{6}{*}{60} & 0.3  & 3593.07 & 3598.29 & 3593.2 & 3599.98 & \textbf{3600*} & 3599.28 & \textbf{3600*}  & \textbf{3600*}  & \textbf{3600*} & 3597.56  \\
  & 0.4  & 3590.68 & 3592.55 & 3591.17 & 3599.97 & 3599.96 & 3598.58 & \textbf{3600*}  & \textbf{3600} & \textbf{3600*}  & 3596.2  \\  
  & 0.5  & 3585.29 & 3585.83 & 3587.5 & 3599.65 & 3599.58 & 3597.53 & 3599.94 & \textbf{3600*} & \textbf{3600*} & 3589.18 \\ 
  & 0.6  & 3572.61 & 3573.7 & 3585.52 & 3595.82 & 3595.85 & 3592.77 & 3596.67 & \textbf{3599.94}  & \textbf{3599.94} & 3578.9 \\ 
  & 0.7  & 3534.71 & 3540.45 & 3561.05 & 3571.47 & 3570.58 & 3566.51 & 3572.12   & 3589.82  & \textbf{3593.52} & 3556.65    \\ 
  & 0.8 & 3478.58 & 3464.14 & 3476.44 & 3478.59 & 3478.37 & 3478.05 & 3478.49  & 3431.85  & 3477.80 & \textbf{3480.03}  \\ 
   \hline
 \multirow{6}{*}{70} & 0.3  & 4890.2 & 4893.75 & 4890.25 & 4899.98 & 4899.98  & 4897.32 & \textbf{4900*}   & \textbf{4900*}   & \textbf{4900*}  & 4896.48  \\
  & 0.4  & 4887.73 & 4888.36  & 4887.98 & 4899.96 & 4899.98 & 4896.4 & 4899.98  & \textbf{4900*} & \textbf{4900*}  & 4887.62 \\  
  & 0.5 & 4881.09 & 4881.17 & 4882.9 & 4899.41 & 4899.44 & 4893.97 & 4899.57   & \textbf{4900*}  & \textbf{4900*} & 4883.67  \\ 
  & 0.6  & 4868.21 & 4868.74 & 4877.77 & 4895.3 & 4894.93 & 4888.52 & 4896.19   & \textbf{4900*} & \textbf{4900*} & 4874.82     \\ 
  & 0.7 & 4829.65 & 4831.94 & 4859.71 & 4872.41 & 4870.97 & 4864.38 & 4872.95  & 4894.58  & \textbf{4896.33} & 4848.31
  \\
  & 0.8  & 4761.44 & 4737.73 & 4761.17 & 4766.67 & 4765.81 & 4763.93 & 4765.91  & 4698.78 & 4766.33 & \textbf{4768.13}  \\
   \hline
\end{tabular}
\label{table:all_plse}

\end{table}    
\end{landscape}

Columns 1 and 2 of Table \ref{all_results_litterature} show the characteristics of each instance (i.e., grid order $n$ and ratio $r$ of pre-assigned symbols). Columns 3-10 present the average number of filled cells in the best solutions obtained by the reference algorithms for the 100 instances of each type $(n,r)$. The results of the proposed MPMA algorithm and Partial-MPMA variant are reported in columns 11 and 12 respectively\footnote{The certificates of the best solutions of MPMA and Partial-MPMA for these 1800 instances are available at \url{https://github.com/GoudetOlivier/MPMA}}. Bold numbers show the dominating values while a star indicates an optimal value (corresponding to the $n^2$ upper bound).

We observe that MPMA always obtain the best scores (in bold) except for the over-constrained instances with $r=0.8$. For the instances with $r=0.8$, the Partial-MPMA variant always obtain the best results.

The best competitors, Tr-ILS* and MMCOL, were launched with a limited amount of available times in \cite{haraguchi2016iterated,jin2019solving}: up to 10 seconds for Tr-ILS* and up to two hours for MMCOL.  In order to verify if these algorithms can improve their results by using more computation time, we ran the codes of these two best performing  algorithms, with a  much  relaxed  time limit of 48 hours. The results are shown in Table \ref{Computation_extend_time}. For each compared algorithm, we report the best and average results over 5 runs ($f_{best}$ and $f_{avg}$) as well as the average computation time needed to reach its best result. 

With this much relaxed time limit, both Tr-ILS* and MMCOL indeed improve their-own results reported in \cite{haraguchi2016iterated} and \cite{jin2019solving} (also shown in Table \ref{table:all_plse}). Meanwhile,  MMCOL and Tr-ILS* are still outperformed by MPMA/Partial-MPMA on the strongly constrained instances with $r \geq 0.7$.

For under-constrained (easy) instances, one notices that MPMA takes much more times to achieve its best results. This comes from the fact that every kernel operation launched on the GPU cannot be stopped until it is completed on each thread. Therefore, even if a solution of the instance is found in one thread, one still needs to wait for all the threads to finish their computation before retrieving the result. In fact, for these easy instances, a very large population with a high diversity is not really mandatory. MPMA can reach the optimal solutions faster with a much reduced population.

\begin{table*}[ht]
\centering
\scriptsize
\caption{Comparison of MPMA/Partial-MPMA with  MMCOL \cite{jin2019solving} and Tr-ILS* \cite{haraguchi2016iterated}
with a much relaxed time limit of 48h on the PLSE instances. \label{Computation_extend_time}}
\scriptsize
\begin{tabular}{ll|lll|lll|lll} 
  \hline
 \multicolumn{2}{c|}{Instance}  &   \multicolumn{3}{|c|}{Tr-ILS* (ext. time)}  &  \multicolumn{3}{|c|}{MMCOL (ext. time)}  &  \multicolumn{3}{|c}{MPMA/Partial-MPMA}  \\
  \hline
    $n$ & $r$  &  $f_{best}$ & $f_{avg}$ &  t(s)  &  $f_{best}$ & $f_{avg}$ &  t(s) & $f_{best}$ & $f_{avg}$ & t(s)  \\
    \hline
  \multirow{7}{*}{50} & 0.3  & \textbf{2500*}  & 2500 & 1 & \textbf{2500*}  & 2500 & 0.22 & \textbf{2500*} & 2500 & 142   \\
  & 0.4  & \textbf{2500*} & 2500 & 2 & \textbf{2500*} & 2500 & 0.16 & \textbf{2500*} & 2500 & 112  \\  
  & 0.5   & \textbf{2500*} & 2500 & 2   & \textbf{2500*} & 2500 & 0.31 & \textbf{2500*} & 2500 & 89  \\ 
  & 0.6   & 2499.63 & 2498.94 & 152   & \textbf{2499.7} & 2499.7 & 17.55 & \textbf{2499.7} & 2499.7 & 489  \\ 
  & 0.7  & 2473.53 & 2472.84 & 511    & 2479.13 & 2478.48 & 46996 & \textbf{2484.38}  & 2483.99
 & 24970 \\ 
  & 0.8 & 2394.34 & 2393.65 & 658   & 2378.15 & 2377.50 & 16268 & \textbf{2394.58} & 2394.42 & 3916 \\
 \hline
  \multirow{6}{*}{60} & 0.3  & \textbf{3600*} & 3600 & 2   & \textbf{3600*} & 3600 & 0.69 & \textbf{3600*} & 3600 & 326  \\
  & 0.4  & \textbf{3600*} & 3600 & 2    & \textbf{3600*} & 3600 & 0.52 & \textbf{3600*} & 3600 & 298  \\  
  & 0.5  & \textbf{3600*} & 3600 & 17   & \textbf{3600*} & 3600 & 0.67 & \textbf{3600*} & 3600 & 214 \\ 
  & 0.6  & \textbf{3599.94} & 3599.25 & 69  & \textbf{3599.94} & 3599.94 & 13.41 & \textbf{3599.94} & 3599.94 & 759  \\ 
  & 0.7  & 3576.7 & 3576.01 & 1388    & 3590.22 & 3589.56 & 49279 & \textbf{3593.52} & 3593.13 &  35658 \\ 
  & 0.8 & 3478.92 & 3478.23 & 460  & 3457.07 & 3456.42 & 77979  & \textbf{3480.05}  & 3479.94 & 18141  \\ 
  \hline
 \multirow{6}{*}{70} & 0.3 & \textbf{4900*} & 4900 & 3   & \textbf{4900*} & 4900 & 0.90 & \textbf{4900*} & 4900 & 721   \\
  & 0.4 & \textbf{4900*} & 4900 & 2  & \textbf{4900*} & 4900 & 0.65 & \textbf{4900*} & 4900 & 489 \\  
  & 0.5   & 4899.71 & 4899.22 & 18    & \textbf{4900*} & 4900 & 1.51 & \textbf{4900*} & 4900 & 349 \\ 
  & 0.6   & 4899.98 & 4899.30 & 437    & \textbf{4900*} & 4900 & 19.82 & \textbf{4900*} & 4900 & 1210  \\ 
  & 0.7  & 4880.10 & 4879.31 & 3245    & 4895.21 & 4894.54 & 55887 & \textbf{4896.33}  & 4895.93 & 46746
  \\
  & 0.8 & 4767.24 & 4766.33 & 2145  & 4736.70 & 4736.07 & 120862 & \textbf{4768.13} & 4767.93 & 56670 \\
 \hline
\end{tabular}
\label{table:ecp_results3}
\end{table*}

On the other hand, using a very large population with a high diversity becomes critical when dealing with the most difficult instances such as those with $r \geq 0.7$. For these instances, MPMA obtains equal or better results compared to Tr-ILS$^*$ and MMCOL for all orders $n=50, 60, 70$. Detailed results for the very difficult instances with $r=0.7$ are displayed in Appendix \ref{app:detailed_results} (Table  \ref{table:detailed_PLSE_results_v1}). Moreover, MPMA can optimally solve 25 of the 100 most challenging instances with $n=70$ and $r=0.7$ (cf. Table \ref{table:detailed_PLSE_results_v1}).

It is difficult to compare the computation time between MPMA and the competitors, as MPMA uses a GPU while the other algorithms use a CPU. Therefore we compare MPMA and MMCOL in terms of number of iterations in order to observe whether the best results of MPMA come from the algorithm itself or from the parallelization. As both MPMA and MMCOL use a one-move tabu search, the number of local search iterations is suitable comparison criterion. We run MPMA and MMCOL with a maximum of $100$ billions iterations of tabu search on the first ten instances of each of the most difficult $(n,r)$ combinations with $n=50, 60, 70$ and $r=70,80$. Each instance is independently solved 5 times. The detailed results are reported in Table \ref{Computation_extend_nb_iterations}, where we show for each instance and each algorithm (MMCOL, MPMA), the best result $f_{best}$ over the 5 trials, the average result $f_{avg}$ over these 5 trials, the average computation time in hours t(h) required to reach the best result and the average number of local search iterations nb\_iter required to reach the best score. The best results are indicated in bold. According to the results, MPMA can achieve better or equal results for all  instances with the same overall number of iterations. In addition, the use of a GPU reduces the time spent by the algorithm, because this important number of iterations can be performed in a shorter amount of time thanks to parallelization. This experiment confirms that the proposed MPMA algorithm dominates MMCOL.

\begin{landscape}
\begin{table*}[ht]
\centering
\scriptsize
\caption{Comparison of MPMA with  MMCOL \cite{jin2019solving} with a large number of iterations on the PLSE instances (maximum of $100 \times 10^9$ iterations). \label{Computation_extend_nb_iterations}}
\begin{tabular}{l|llll|llll||l|llll|llll} 
  \hline
 Instance   &  \multicolumn{4}{|c|}{MMCOL (ext. nb iter.)}  &  \multicolumn{4}{|c||}{MPMA} & Instance   &  \multicolumn{4}{|c|}{MMCOL (ext. nb iter)}  &  \multicolumn{4}{|c}{MPMA} \\
  \hline
      &  $f_{best}$ & $f_{avg}$ &  t(h) & nb\_iter. & $f_{best}$ & $f_{avg}$ & t(h) & nb\_iter  &  &  $f_{best}$ & $f_{avg}$ &  t(h) & nb\_iter & $f_{best}$ & $f_{avg}$ & t(h) & nb\_iter \\
    \hline
QC-50-70-1 & 2479  & 2478.2 & 168 & $82 \times 10^9$ & \textbf{2485} & 2484.0 & 4 & $36 \times 10^9$ & QC-50-80-1 &  2380 & 2379.6 & 38 & $14 \times 10^9$ & \textbf{2393} & 2392.8 & 0.2 &  $3 \times 10^9$ \\ 
QC-50-70-2 & 2477  & 2476.6 & 45 & $24 \times 10^9$ & \textbf{2482} & 2482.0 & 2 & $20 \times 10^9$ & QC-50-80-2 & 2377  & 2376.4 & 18 & $9 \times 10^9$ & \textbf{2391} &  2391 & 0.03 & $0.5 \times 10^9$ \\ 
QC-50-70-3 & 2487  & 2486.6 & 150 & $81 \times 10^9$ & \textbf{2490}  & 2489.6 & 5 & $44 \times 10^9$ & QC-50-80-3 &  2381 & 2380.2 & 3 & $1 \times 10^9$ & \textbf{2395} & 2395 & 0.5 &  $7 \times 10^9$\\ 
QC-50-70-4 & 2482  & 2481 & 54 & $29 \times 10^9$ & \textbf{2487}  & 2486.8 & 3 & $28 \times 10^9$ & QC-50-80-4 &  2386 & 2385.6 & 15 & $6 \times 10^9$ &  \textbf{2399} & 2399 & 0.03 & $0.5 \times 10^9$  \\ 
QC-50-70-5 &  2474 & 2474 & 139 & $68 \times 10^9$ & \textbf{2482} & 2481.6 & 10 & $94 \times 10^9$ & QC-50-80-5 & 2377  & 2376.4 & 15 & $6 \times 10^9$ & \textbf{2388} & 2388 & 1.5 & $27 \times 10^9$  \\ 
QC-50-70-6 & 2481  & 2480.2 & 28 & $14 \times 10^9$ & \textbf{2485}  & 2484.6  & 5 & $44 \times 10^9$ & QC-50-80-6 &  2378 & 2377.8 & 5 & $2 \times 10^9$ & \textbf{2393} & 2393  & 2 &  $34 \times 10^9$ \\ 
QC-50-70-7 & 2480  & 2479.6 & 186 & $92 \times 10^9$ & \textbf{2485}  & 2485.0 & 6 & $57 \times 10^9$ & QC-50-80-7 & 2387  & 2387 & 121 & $44 \times 10^9$ & \textbf{2404} & 2403.8 & 0.2  &  $3 \times 10^9$ \\ 
QC-50-70-8 & 2476  & 2475.4 & 98 & $49 \times 10^9$ & \textbf{2483} & 2482.6 & 6  & $62 \times 10^9$ & QC-50-80-8 &  2367 & 2366.4 & 38 & $14 \times 10^9$ & \textbf{2389} & 2389 & 0.4 & $6 \times 10^9$  \\ 
QC-50-70-9 & 2483  & 2482.4 & 45  & $24 \times 10^9$ & \textbf{2486}  & 2485.8 & 8 & $84 \times 10^9$ & QC-50-80-9 &  2378 & 2377.2 & 2 & $2 \times 10^9$ & \textbf{2393} & 2393 &  0.2 &  $4 \times 10^9$\\ 
QC-50-70-10 &  2472 & 2471.4 & 26 & $10 \times 10^9$ & \textbf{2480} & 2479.6 & 9  & $75 \times 10^9$ & QC-50-80-10 &  2362 & 2361 & 45 & $11 \times 10^9$ & \textbf{2382} & 2382  & 0.03  & $0.5 \times 10^9$ \\ 
 \hline
QC-60-70-1 & 3593  & 3592.8 & 121 & $51 \times 10^9$ & \textbf{3594} & 3593.8 & 6 & $39 \times 10^9$ & QC-60-80-1 & 3448  & 3449.2 & 147  & $27 \times 10^9$ & \textbf{3467} & 3467 & 0.2 & $2 \times 10^9$ \\ 
QC-60-70-2 & 3590  & 3589.4 & 71 & $36 \times 10^9$ & \textbf{3594} & 3593.2 & 9 & $59 \times 10^9$ & QC-60-80-2 &  3453 & 3452.2 & 88 & $20 \times 10^9$ & \textbf{3472} & 3472 & 3  &  $22 \times 10^9$ \\
QC-60-70-3 &  3578 & 3577.2 & 160 & $55 \times 10^9$ & \textbf{3583} & 3582.4 & 14  & $91 \times 10^9$ & QC-60-80-3 & 3454  & 3452.6 & 50 & $11 \times 10^9$ & \textbf{3475} & 3474.8 & 2  &  $21 \times 10^9$ \\ 
QC-60-70-4 & 3592  & 3591 &  164 & $58 \times 10^9$ & \textbf{3595} & 3595 & 11 & $72 \times 10^9$ & QC-60-80-4 & 3464  &  3463.2 & 50 & $10 \times 10^9$ & \textbf{3482} & 3482 & 1  &  $10 \times 10^9$ \\ 
QC-60-70-5 &  3592 & 3591.2 & 267 & $94 \times 10^9$ & \textbf{3594} & 3593.8 &  7 & $42 \times 10^9$ & QC-60-80-5 &  3471 & 3470.4 & 73 & $15 \times 10^9$ & \textbf{3490} & 3489.6 & 3 & $40 \times 10^9$ \\ 
QC-60-70-6 & 3596  & 3595  & 287 & $99 \times 10^9$ & \textbf{3598} & 3597.0 & 10 & $64 \times 10^9$ & QC-60-80-6 & 3450  &   3448 & 233 & $45 \times 10^9$ & \textbf{3476} & 3476 & 0.5  & $5 \times 10^9$ \\ 
QC-60-70-7 & 3589  & 3588 &  162 & $57 \times 10^9$ & \textbf{3591} & 3590.8 & 8 & $52 \times 10^9$ & QC-60-80-7 & 3459  & 3458.4 & 75 & $31 \times 10^9$ & \textbf{3478} & 3478 & 1.5 & $16 \times 10^9$ \\ 
QC-60-70-8 &  3590 & 3589.2 & 241 & $85 \times 10^9$ & \textbf{3593} & 3592.6 &  9 & $59 \times 10^9$ & QC-60-80-8 &  3464 & 3462.8 & 49 & $24 \times 10^9$ & \textbf{3488} &  3488 & 0.3 & $3 \times 10^9$ \\ 
QC-60-70-9 & 3592  & 3592 &  94  & $36 \times 10^9$ & \textbf{3595} & 3594.4 & 13 & $85 \times 10^9$ & QC-60-80-9 &  3448 & 3447.6 & 53 & $18 \times 10^9$ & \textbf{3471} & 3471 & 1.5 & $16 \times 10^9$ \\ 
QC-60-70-10 &  3590 &  3589.2 & 235 & $65 \times 10^9$ & \textbf{3592} & 3591.4 & 8 & $51 \times 10^9$ & QC-60-80-10 & 3457  & 3456.6 & 88 & $26 \times 10^9$ & \textbf{3477} & 3477 & 1.4 & $15 \times 10^9$ \\ 
  \hline
QC-70-70-1 & 4895  & 4894.8  & 47 & $15 \times 10^9$ & \textbf{4897} & 4897.0 & 18 & $84 \times 10^9$ & QC-70-80-1 & 4729  & 4727.2 & 59 & $9 \times 10^9$ & \textbf{4766} & 4764.8 & 2 & $16 \times 10^9$ \\ 
QC-70-70-2 & 4895  &  4895 & 151 & $51 \times 10^9$ & \textbf{4896} & 4896.0 & 9 & $42 \times 10^9$ & QC-70-80-2 &  4738 & 4736.4 & 23 & $4 \times 10^9$ & \textbf{4771} & 4768.6 & 4 & $33 \times 10^9$ \\ 
QC-70-70-3 & 4897  &  4896.6 & 125 & $48 \times 10^9$ & 4897 & 4897.0 & 11 & $51 \times 10^9$ & QC-70-80-3 &  4721 & 4720.4 & 22 & $3 \times 10^9$ & \textbf{4756} & 4755 & 4 & $28 \times 10^9$ \\ 
QC-70-70-4 &  4893 & 4892.4  & 69 & $27 \times 10^9$ & \textbf{4894} & 4893.8 & 17 & $79 \times 10^9$  & QC-70-80-4 &  4735 & 4733.4 & 124 & $22 \times 10^9$ & \textbf{4770}  & 4767.8 & 2 & $17 \times 10^9$ \\ 
QC-70-70-5 & 4898  &  4898 & 177 & $91 \times 10^9$ & 4898 & 4897.6 & 8 & $38 \times 10^9$ & QC-70-80-5 & 4748  & 4745.8 & 174 & $32 \times 10^9$ & \textbf{4768}  & 4768 & 2 & $19 \times 10^9$ \\ 
QC-70-70-6 &  4900 & 4898.6  & 63 & $29 \times 10^9$ & 4900 & 4899.2 & 7 & $32 \times 10^9$ & QC-70-80-6 & 4747  & 4746.2 & 82 & $14 \times 10^9$ & \textbf{4773} & 4772.8 & 3 & $24 \times 10^9$ \\ 
QC-70-70-7 & 4898  & 4897.8  & 146 & $71 \times 10^9$ & 4898 & 4897.6 & 19 & $89 \times 10^9$ & QC-70-80-7 &  4746 & 4744 & 19 & $3 \times 10^9$ & \textbf{4774} & 4774 & 1 & $9 \times 10^9$ \\ 
QC-70-70-8 & 4897 &  4896.6 & 189 & $92 \times 10^9$ & \textbf{4898} &  4898 & 9 & $43 \times 10^9$ & QC-70-80-8 &  4745 & 4744.2 & 57 & $8 \times 10^9$ & \textbf{4777} & 4775 & 2 & $15 \times 10^9$ \\ 
QC-70-70-9 & 4897  &  4896.2 & 85 & $52 \times 10^9$ & \textbf{4898} & 4896.8 & 16 & $81 \times 10^9$ & QC-70-80-9 & 4730  & 4728.8 & 8 & $1 \times 10^9$ & \textbf{4761} & 4760.4 & 2 & $15 \times 10^9$ \\ 
QC-70-70-10 & 4897  &  4896.4 & 93 & $46 \times 10^9$ & 4897 & 4896.8 & 7 & $31 \times 10^9$ & QC-70-80-10 & 4750  & 4748.8 & 60 & $9 \times 10^9$ & \textbf{4777} & 4776 & 4 & $28 \times 10^9$ \\ 
\hline
\end{tabular}
\label{table:ecp_results3}
\end{table*} 
\end{landscape}

In summary, MPMA and its Partial-MPMA variant for highly constrained instances (when $r > 0.7$) compete very favorably with the best performing PLSE methods in the literature, by reporting equal or better results on the 1800 benchmark instances. In Appendix \ref{sec:results_LSC}, we show that MPMA also performs extremely well on the special case of the Latin square completion problem, by attaining the optimal solutions for all the LSC benchmark instances.

\section{Analysis of important factors in the algorithm \label{sec:key_components}} 

In this section, we analyze the impacts of three important factors of the MPMA algorithm: (i) its very large population, the AUX crossover and (iii) the nearest neighbor matching strategy for parent selection. These experiments are based on the first ten hard instances with $n=60$ and $r=0.7$ of the PLSE. 

\subsection{Sensitivity to the population size}

\begin{figure}[!ht]
    \centering
    \includegraphics[width=1.0\textwidth]{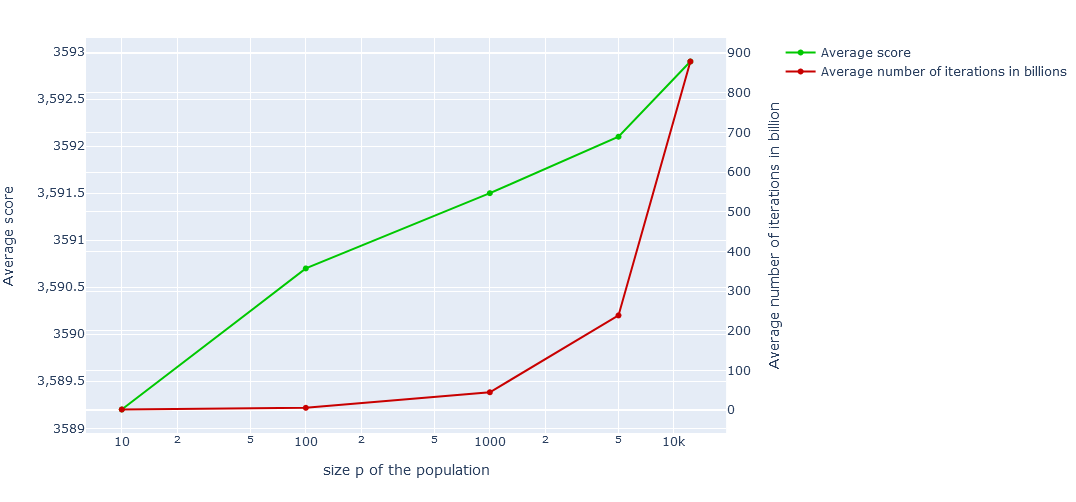}
    \caption{Impact of the population size $p$ on the performance of MPMA. Green curve corresponds to the average score and red curve to the average number of iterations in billions required to reach the best scores.}\label{fig:sensi_pop} 
\end{figure}

We first perform a sensitivity analysis of the algorithm with respect to the population size. For this, we perform the MPMA algorithm with $p$ varying in the range $[10,12288]$ to solve each of the ten instance 5 times under a time limit to 20 hours per run. Figure \ref{fig:sensi_pop} displays the sensitivity of the average results to the population size $p$.

For the same time budget, the MPMA algorithm obtains better results with a larger size. When $p = 12288$, the algorithm always attains the best score over 10 runs. This can be explained by two reasons.  First, due to the parallelization of the calculations on the GPUs, a large population improves the diversity of the population and helps the algorithm to perform a higher average global number of iterations at each run, which in turn increases the chance to attain high-quality solutions. Second, a large population increases the chance for each individual to find a closer but different nearest neighbor in the population for parent matching of the AUX crossover, which helps to generate promising offspring solutions.

\subsection{Impact of the asymmetric uniform crossover }

To study the impact of the asymmetric uniform crossover AUX on the MPMA algorithm, we compare it with four different variants of MPMA where the AUX crossover described in Section \ref{sec:AUXcross} is changed or disabled.
\begin{itemize}
\item The first variant is a baseline variant without crossover, so each offspring is an exact copy of its first parent.
\item The greedy partition crossover GPX \cite{galinier1999hybrid} is adapted for the Latin square problem: each color class of the offspring inherits the largest color class of the selected parent.
\item The AUX crossover is replaced by the maximum approximate group based crossover MAGX of the MMCOL algorithm for the related Latin square completion problem \cite{jin2019solving}.
\item The AUX crossover is replaced by the uniform crossover (UX) which corresponds to AUX with $p_{ij}$ being fixed to the value of $0.5$.
\end{itemize}

\begin{figure}[!ht]
    \centering
    \includegraphics[width=1.0\textwidth]{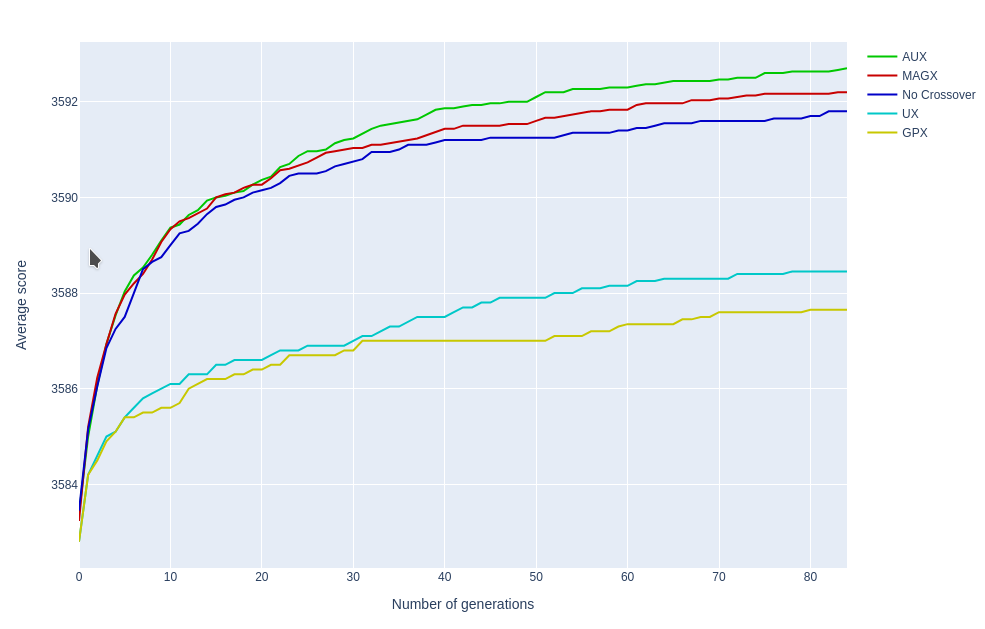}
    \caption{Comparison of five different MPMA variants: No crossover (blue), GPX (yellow), MAGX (red), UX (light blue), AUX (green).}\label{fig:crossovers} 
\end{figure}

Figure \ref{fig:crossovers} shows the evolution of the best fitness values averaged over 5 runs for the same ten PLSE instances with $(n,r)=(60,0.7)$ through the number of generations of each algorithm. One notices that the crossovers GPX and UX, which are the most disruptive, perform badly and are even outperformed by the variant without crossovers (blue line). This can be explained by the fact that the individuals are very distant in the population and rarely share large common features. Indeed, we experimentally observed that the average pairwise distance in the population is usually very large, around $0.7 \times |V|$. 

The AUX and MAGX crossovers perform the best and dominate GPX and UX.  Meanwhile, AUX dominates  MAGX after 50 generations in average. The difference is statistically significant (confirmed by t-test with the p-value of 0.001). One reason to explain the advantage of AUX over MAGX is that with the AUX crossover, the offspring inherits more features from one parent than from the other parent. On the contrary, since MAGX is a symmetric crossover, crossing-over $(S_i,S_j)$ and $(S_j,S_i)$ lead to the same offspring, which results in less diversified offspring in the next generation.

\subsection{Impact of the crossover matching strategy} 

To study the impact of the nearest neighbor matching strategy for the AUX crossover, we run a MPMA variant where this matching strategy is replaced by a random matching strategy: each individual as the first parent is cross-overed with another individual chosen randomly in the population. 

\begin{figure}[!ht]
    \centering
    \includegraphics[width=0.85\textwidth]{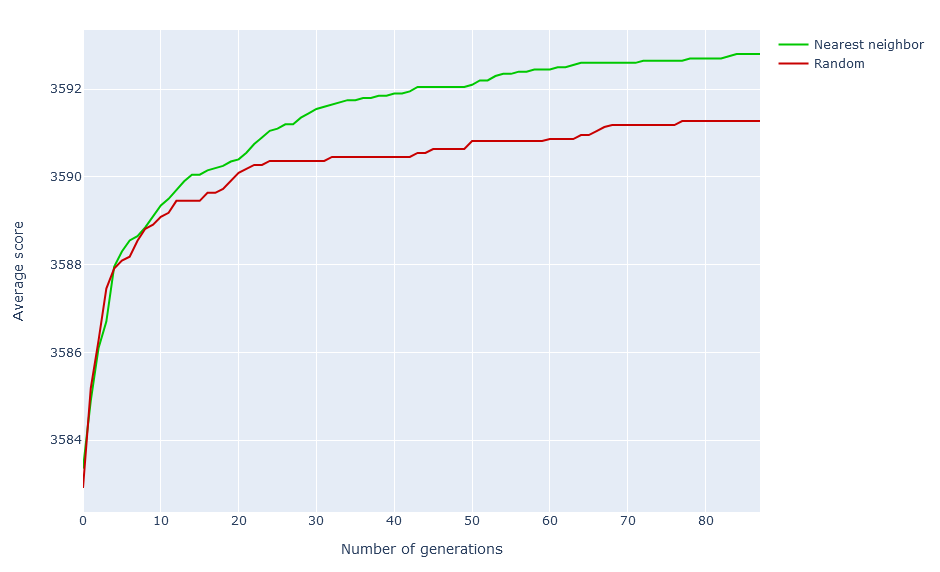}
    \caption{Comparison of two parent matching strategies in MPMA: random matching (red) and nearest neighbor matching (green).}\label{fig:matching} 
\end{figure}

Figure \ref{fig:matching} shows the evolution of the best fitness values averaged over 5 runs for the same 10 first PLSE instances with $(n,r)=(60,0.7)$ with respect to the number of generations of the algorithm. 
One notices that the matching strategy has an important impact on the performance. The dominance of the nearest neighbor matching strategy over the random matching becomes more and more evident after 10 generations. The difference is statistically significant (t-test with the p-value of 0.001). This is because two parents chosen randomly in the very large population share little information, leading to poor offspring whose quality can be hardly raised even after local optimization. The nearest neighbor strategy avoids this problem, as it does not destroy too much the color classes transmitted to the offspring, while preserving a certain level of diversity. This creates opportunities for the subsequent local search to explore new and interesting areas of the search space. 

\section{Conclusion \label{sec:conclusion}} 

We presented a massively parallel population-based algorithm with a very large population and a practical implementation on GPUs to solve the partial Latin square extension problem as well as the special case of the Latin square completion problem. This approach highlights the interest of a very large population that enables massively parallel local optimization, offspring generations and distance calculations. The algorithm features a parameterized asymmetric crossover equipped with a dedicated parent matching strategy to build promising offspring, an effective parallel two-phase tabu search to improve new solutions and an original pool updating mechanism. 

We performed extensive experiments to assess the proposed algorithm on the set of 1800 benchmark instances with various orders and ratios of pre-filled cells. The results showed that the algorithm obtains state-of-the-art results in average for all Latin square configurations  $(n,r)$. Furthermore, it definitely closed 25 challenging instances of order $n=70$ and ratio $r=0.7$. We investigated the impacts of key algorithmic components including the large population size and the parent matching strategy. This work demonstrates for the first time the high potential of GPU-based parallel computations for solving the challenging Latin square extension problem, by exploiting the formidable computing power offered by the GPUs and designing suitable search strategies.

The proposed algorithm can be used to solve relevant problems related to the PLSE. The availability of the source code of our algorithm will facilitate such applications. The design ideas of the algorithm can help to develop effective algorithms for other difficult combinatorial optimization problems. Future works could be carried out in particular to improve the parent matching strategy. For instance, it would be interesting to investigate strategies driven by a deep graph convolutional neural network in order to build the most promising offspring from appropriate parents. 

\section*{CRediT author statement}

\textbf{Olivier Goudet:} Conceptualization, Methodology, Software, Investigation,  Writing - Original Draft. \textbf{Jin-Kao Hao:} Conceptualization, Methodology, Investigation,  Writing - Original Draft.

\section*{Acknowledgment}

We would like to thank Dr. Kazuya Haraguchi for sharing his Tr-ILS$^*$ code and the instances \cite{haraguchi2016iterated}. We are grateful to Dr. Yan Jin for her assistance on their MMCOL code \cite{jin2019solving}.  This work was granted access to the HPC resources of IDRIS (Grant No. 2020-A0090611887) from GENCI.

\bibliographystyle{elsarticle-num} 
\bibliography{biblio}

\appendix
\section{Detailed results for the challenging PLSE instances with $r=0.7$ \label{app:detailed_results}}

According to \cite{haraguchi2016iterated}, instances with $r=0.7$ are among the most challenging. Table \ref{table:detailed_PLSE_results_v1}  present the detailed results obtained by the MPMA algorithm on the three sets of 300 PLSE instance with $r=0.7$ and $n=50,60,70$. Column 1 identifies the instances of each type ($n,r$). For each instance, we report the best PLSE score $f_{best}$ (i.e., the largest number of filled cells) obtained over 5 runs with a maximum of 100 billions of tabu iterations, average score $f_{avg}$ and average computation time t(s) in seconds to reach the best results. Bold values are the record-breaking results compared to the best-known results in the literature (including the best results obtained by running the codes of Tr-ILS$^*$ \cite{haraguchi2016iterated} and MMCOL \cite{jin2019solving} with the extended time limit of 48h). A star indicates an optimal value. The optimality is proved if (i) the number of filled cells reaches the upper bound $n^2 - l$ if $l\neq 1$ (cf. Section \ref{preprocess_Latin_square}), or (ii) the number of filled cells is $n^2 - 2$ if $l=1$ (cf. Theorem 6 in \cite{donovan2000completion}). One observes that MPMA improves the best-known results for a large majority of the 300 instances and closes definitively 25 instances by reaching their optimal scores. Among these 25 optimal results, 14 were also achieved by MMCOL (starred non-bold values) with the extended time limit.

\renewcommand{\baselinestretch}{0.6}\huge\normalsize

\begin{table}[H]
\centering
\tiny
\caption{Detailed results of MPMA  for the PLSE instance with $r=0.7$}
\begin{tabular}{l|l @{\hspace{0.2cm}} l @{\hspace{0.4cm}} l||l @{\hspace{0.2cm}} l @{\hspace{0.4cm}} l||l @{\hspace{0.2cm}} l @{\hspace{0.4cm}} l} 
   \hline
   &   \multicolumn{3}{|c|}{PLSE-50-70} &  \multicolumn{3}{|c||}{PLSE-60-70}   & \multicolumn{3}{|c|}{PLSE-70-70}  \\
   \hline
    $Id$  & $f_{best}$ & $f_{avg}$ &  t(s)  & $f_{best}$ & $f_{avg}$ & t(s) &  $f_{best}$ & $f_{avg}$ &  t(s)   \\
    \hline
1 & \textbf{2485} & 2484.0 & 14634 & \textbf{3594} & 3593.8 & 21133 & \textbf{4897} & 4897.0 & 64501 \\
2 & \textbf{2482} & 2482.0 & 7979 & \textbf{3594} & 3593.2 & 32775 & \textbf{4896} & 4896.0 & 31215 \\
3 & \textbf{2490} & 2489.6 & 16640 & \textbf{3583} & 3582.4 & 50463 & 4897 & 4897.0 & 38822 \\
4 & \textbf{2487} & 2486.8 & 11040 & \textbf{3595} & 3595.0 & 42003 & \textbf{4894} & 4893.8 & 60959 \\
5 & \textbf{2482} & 2481.6 & 38581 & \textbf{3594} & 3593.8 & 23419 & 4898* & 4897.6 & 29179 \\
6 & \textbf{2485} & 2484.6 & 17336 & \textbf{3598} & 3597.0 & 35415 & 4900* & 4899.2 & 24546 \\
7 & \textbf{2485} & 2485.0 & 21093 & \textbf{3591} & 3590.8 & 29223 & 4898* & 4897.6 & 68794 \\
8 & \textbf{2483} & 2482.6 & 22696 & \textbf{3593} & 3592.6 & 31549 & \textbf{4898} & 4898.0 & 34353 \\
9 & \textbf{2486} & 2485.8 & 29292 & \textbf{3595} & 3594.4 & 45923 & \textbf{4898}* & 4896.8 & 56775 \\
10 & \textbf{2480} & 2479.6 & 33675 & \textbf{3592} & 3591.4 & 27668 & 4897 & 4896.8 & 24600 \\
11 & \textbf{2488} & 2488.0 & 10494 & \textbf{3591} & 3591.0 & 43547 & \textbf{4895} & 4895.0 & 41911 \\
12 & \textbf{2485} & 2484.8 & 11099 & \textbf{3595} & 3595.0 & 29279 & \textbf{4895} & 4895.0 & 49769 \\
13 & \textbf{2483} & 2482.0 & 35398 & \textbf{3591} & 3590.6 & 30085 & 4896 & 4896.0 & 38843 \\
14 & \textbf{2483} & 2483.0 & 24327 & 3596 & 3594.8 & 12871 & 4900* & 4900.0 & 49655 \\
15 & \textbf{2483} & 2483.0 & 22104 & \textbf{3598} & 3597.6 & 42935 & \textbf{4897}* & 4897.0 & 44162 \\
16 & \textbf{2484} & 2484.0 & 24908 & \textbf{3589} & 3588.4 & 46321 & \textbf{4895} & 4894.4 & 51131 \\
17 & \textbf{2486} & 2486.0 & 30868 & \textbf{3589} & 3588.2 & 44392 & \textbf{4898}* & 4897.8 & 55798 \\
18 & \textbf{2489} & 2488.6 & 43310 & \textbf{3594} & 3594.0 & 31095 & \textbf{4896} & 4895.2 & 45508 \\
19 & \textbf{2485} & 2485.0 & 46223 & \textbf{3592} & 3591.6 & 34286 & \textbf{4896} & 4895.0 & 33190 \\
20 & \textbf{2490} & 2490.0 & 66822 & \textbf{3595} & 3595.0 & 45880 & 4898* & 4898.0 & 45274 \\
21 & \textbf{2483} & 2482.8 & 10055 & \textbf{3594} & 3593.8 & 28887 & \textbf{4896} & 4895.8 & 43046 \\
22 & \textbf{2484} & 2483.8 & 31473 & \textbf{3594} & 3593.6 & 36060 & \textbf{4892} & 4891.8 & 51100 \\
23 & \textbf{2485} & 2485.0 & 59471 & \textbf{3595} & 3594.8 & 35139 & 4898* & 4897.4 & 57851 \\
24 & \textbf{2488} & 2487.2 & 39261 & \textbf{3595} & 3594.2 & 39917 & \textbf{4896} & 4895.6 & 62074 \\
25 & \textbf{2484} & 2484.0 & 67246 & \textbf{3595} & 3595.0 & 24632 & \textbf{4896} & 4895.6 & 29744 \\
26 & \textbf{2483} & 2482.4 & 11660 & \textbf{3591} & 3590.4 & 29959 & \textbf{4896} & 4896.0 & 49052 \\
27 & \textbf{2481} & 2480.8 & 17530 & \textbf{3596} & 3595.6 & 21738 & \textbf{4896} & 4895.2 & 64276 \\
28 & \textbf{2484} & 2484.0 & 57286 & \textbf{3593} & 3592.2 & 39360 & \textbf{4895} & 4895.0 & 30988 \\
29 & \textbf{2486} & 2486.0 & 24712 & \textbf{3594} & 3593.8 & 36996 & \textbf{4894} & 4893.0 & 33507 \\
30 & \textbf{2485} & 2484.4 & 32252 & \textbf{3594} & 3593.2 & 29165 & \textbf{4894} & 4893.8 & 67407 \\
31 & \textbf{2481} & 2480.2 & 30863 & \textbf{3596} & 3595.8 & 28915 & \textbf{4895} & 4894.8 & 66784 \\
32 & \textbf{2481} & 2480.8 & 26209 & \textbf{3598} & 3597.8 & 36874 & \textbf{4898}* & 4898.0 & 63561 \\
33 & \textbf{2483} & 2482.0 & 15300 & \textbf{3594} & 3593.8 & 50209 & \textbf{4893} & 4892.8 & 53578 \\
34 & \textbf{2484} & 2483.6 & 17261 & \textbf{3595} & 3594.8 & 25322 & \textbf{4896} & 4896.0 & 29581 \\
35 & \textbf{2483} & 2482.2 & 9258 & \textbf{3594} & 3593.6 & 48250 & \textbf{4895} & 4893.8 & 37046 \\
36 & \textbf{2484} & 2483.8 & 39607 & \textbf{3589} & 3588.8 & 52302 & \textbf{4896} & 4896.0 & 30549 \\
37 & \textbf{2486} & 2486.0 & 30891 & \textbf{3592} & 3591.8 & 36935 & 4898 & 4897.4 & 33335 \\
38 & \textbf{2479} & 2479.0 & 27487 & \textbf{3593} & 3593.0 & 42965 & \textbf{4896} & 4895.8 & 46231 \\
39 & \textbf{2482} & 2482.0 & 17885 & \textbf{3592} & 3592.0 & 40127 & \textbf{4895} & 4895.0 & 45687 \\
40 & \textbf{2486} & 2485.8 & 25149 & \textbf{3584} & 3584.0 & 28671 & \textbf{4897} & 4897.0 & 39611 \\
41 & \textbf{2486} & 2484.8 & 20498 & \textbf{3593} & 3593.0 & 44563 & \textbf{4894} & 4894.0 & 68759 \\
42 & \textbf{2485} & 2484.2 & 29963 & \textbf{3596} & 3594.8 & 20232 & 4900* & 4900.0 & 37155 \\
43 & \textbf{2486} & 2485.6 & 22424 & \textbf{3592} & 3591.8 & 33863 & \textbf{4897} & 4896.4 & 65871 \\
44 & \textbf{2478} & 2478.0 & 21238 & \textbf{3596} & 3595.2 & 39637 & 4900* & 4900.0 & 24920 \\
45 & \textbf{2487} & 2486.2 & 4387 & \textbf{3594} & 3593.2 & 53331 & \textbf{4896} & 4895.6 & 36570 \\
46 & \textbf{2486} & 2485.2 & 8202 & \textbf{3590} & 3590.0 & 31509 & \textbf{4895} & 4894.4 & 66643 \\
47 & \textbf{2483} & 2483.0 & 15598 & \textbf{3596} & 3596.0 & 50515 & \textbf{4896} & 4895.2 & 28201 \\
48 & \textbf{2485} & 2485.0 & 12008 & \textbf{3594} & 3594.0 & 52813 & \textbf{4898}* & 4898.0 & 45563 \\
49 & \textbf{2488} & 2487.8 & 19546 & \textbf{3592} & 3592.0 & 30852 & \textbf{4896} & 4896.0 & 61217 \\
50 & \textbf{2487} & 2486.2 & 36084 & \textbf{3591} & 3590.4 & 28170 & \textbf{4892} & 4891.2 & 63410 \\
51 & \textbf{2482} & 2482.0 & 14454 & \textbf{3597} & 3596.8 & 27863 & \textbf{4894} & 4893.2 & 52307 \\
52 & \textbf{2483} & 2482.8 & 3734 & \textbf{3594} & 3593.2 & 29788 & \textbf{4894} & 4893.6 & 47142 \\
53 & \textbf{2479} & 2478.2 & 29808 & \textbf{3590} & 3590.0 & 34304 & \textbf{4895} & 4895.0 & 61063 \\
54 & \textbf{2482} & 2482.0 & 31105 & \textbf{3595} & 3595.0 & 36915 & 4895 & 4894.8 & 49282 \\
55 & \textbf{2490} & 2490.0 & 57119 & \textbf{3593} & 3592.8 & 43977 & 4898 & 4898.0 & 42535 \\
56 & \textbf{2486} & 2485.2 & 16890 & \textbf{3594} & 3594.0 & 26958 & \textbf{4897} & 4896.6 & 40521 \\
57 & \textbf{2485} & 2484.0 & 17693 & \textbf{3596} & 3595.6 & 22850 & \textbf{4897} & 4895.8 & 36423 \\
58 & \textbf{2484} & 2483.6 & 22020 & \textbf{3592} & 3590.8 & 42025 & \textbf{4895} & 4895.0 & 50358 \\
59 & \textbf{2479} & 2479.0 & 17566 & \textbf{3597} & 3597.0 & 35690 & \textbf{4897} & 4896.2 & 48762 \\
60 & \textbf{2485} & 2483.8 & 12812 & \textbf{3594} & 3594.0 & 49378 & 4898* & 4897.6 & 41952 \\
61 & \textbf{2488} & 2487.6 & 32457 & \textbf{3593} & 3593.0 & 34521 & \textbf{4896} & 4895.6 & 40522 \\
62 & \textbf{2483} & 2482.2 & 11236 & \textbf{3595} & 3595.0 & 36297 & \textbf{4897} & 4896.6 & 26971 \\
63 & \textbf{2484} & 2483.8 & 49638 & \textbf{3593} & 3593.0 & 33612 & 4895 & 4894.0 & 36138 \\
64 & \textbf{2487} & 2486.8 & 18411 & \textbf{3589} & 3589.0 & 45479 & \textbf{4896} & 4895.8 & 43970 \\
65 & \textbf{2483} & 2483.0 & 14955 & \textbf{3594} & 3592.8 & 51097 & \textbf{4895} & 4894.2 & 64374 \\
66 & \textbf{2487} & 2486.0 & 6173 & \textbf{3594} & 3593.8 & 32932 & \textbf{4897} & 4895.8 & 29134 \\
67 & \textbf{2492} & 2491.4 & 13935 & \textbf{3596} & 3594.8 & 46629 & \textbf{4896} & 4895.2 & 39470 \\
68 & \textbf{2485} & 2484.6 & 13185 & \textbf{3591} & 3591.0 & 46103 & \textbf{4894} & 4893.6 & 65397 \\
69 & \textbf{2480} & 2478.8 & 61028 & \textbf{3597} & 3596.8 & 29333 & \textbf{4896} & 4895.4 & 33751 \\
70 & \textbf{2480} & 2480.0 & 10097 & \textbf{3596} & 3595.0 & 21332 & \textbf{4897} & 4896.4 & 70332 \\
71 & \textbf{2485} & 2484.8 & 14403 & \textbf{3597} & 3596.8 & 26326 & \textbf{4898}* & 4896.8 & 37049 \\
72 & \textbf{2485} & 2485.0 & 23233 & \textbf{3593} & 3593.0 & 41770 & \textbf{4898}* & 4897.6 & 55525 \\
73 & \textbf{2481} & 2481.0 & 13541 & \textbf{3592} & 3591.8 & 42388 & \textbf{4897} & 4896.2 & 41423 \\
74 & \textbf{2487} & 2486.2 & 17062 & \textbf{3591} & 3590.4 & 41805 & \textbf{4895} & 4894.8 & 66514 \\
75 & \textbf{2486} & 2485.8 & 6442 & \textbf{3591} & 3589.8 & 37734 & \textbf{4893} & 4892.6 & 27458 \\
76 & \textbf{2482} & 2481.6 & 31480 & \textbf{3596} & 3595.6 & 32075 & 4895 & 4894.4 & 37566 \\
77 & \textbf{2484} & 2484.0 & 30772 & \textbf{3594} & 3593.6 & 30518 & \textbf{4898}* & 4897.8 & 71196 \\
78 & \textbf{2485} & 2484.2 & 25027 & \textbf{3594} & 3593.8 & 37885 & \textbf{4894} & 4892.8 & 61918 \\
79 & \textbf{2486} & 2485.4 & 11737 & \textbf{3592} & 3592.0 & 13140 & \textbf{4895} & 4894.8 & 55482 \\
80 & \textbf{2486} & 2485.0 & 11477 & \textbf{3591} & 3590.6 & 17375 & \textbf{4895} & 4893.8 & 43852 \\
81 & \textbf{2484} & 2483.6 & 25867 & \textbf{3594} & 3594.0 & 49588 & \textbf{4898} & 4898.0 & 34218 \\
82 & \textbf{2486} & 2486.0 & 12979 & \textbf{3596} & 3595.2 & 49521 & \textbf{4896} & 4895.0 & 54607 \\
83 & \textbf{2484} & 2483.8 & 45426 & 3591 & 3591.0 & 36875 & \textbf{4897} & 4896.8 & 53219 \\
84 & \textbf{2482} & 2480.8 & 13416 & \textbf{3597} & 3596.8 & 39570 & \textbf{4896} & 4895.6 & 55558 \\
85 & \textbf{2486} & 2485.2 & 18071 & \textbf{3591} & 3591.0 & 54486 & \textbf{4895} & 4894.2 & 51972 \\
86 & \textbf{2484} & 2483.6 & 21323 & \textbf{3591} & 3590.8 & 32150 & \textbf{4896} & 4895.2 & 64547 \\
87 & \textbf{2482} & 2481.8 & 7322 & \textbf{3597} & 3596.2 & 21235 & \textbf{4898} & 4897.8 & 58645 \\
88 & \textbf{2486} & 2484.8 & 54905 & \textbf{3595} & 3594.0 & 31176 & 4898* & 4897.6 & 55316 \\
89 & \textbf{2483} & 2483.0 & 22200 & \textbf{3596} & 3595.2 & 42334 & 4898* & 4898.0 & 69525 \\
90 & \textbf{2484} & 2483.4 & 41303 & \textbf{3591} & 3590.6 & 28078 & 4898* & 4897.8 & 33387 \\
91 & \textbf{2485} & 2485.0 & 27282 & \textbf{3595} & 3594.8 & 37660 & \textbf{4898}* & 4897.8 & 35507 \\
92 & \textbf{2480} & 2478.8 & 64358 & \textbf{3595} & 3595.0 & 20591 & \textbf{4898} & 4898.0 & 44032 \\
93 & \textbf{2485} & 2484.6 & 54895 & \textbf{3595} & 3594.4 & 34376 & 4898* & 4897.4 & 30656 \\
94 & \textbf{2488} & 2487.4 & 36221 & \textbf{3593} & 3592.6 & 27382 & \textbf{4898}* & 4898.0 & 42838 \\
95 & \textbf{2485} & 2484.2 & 34930 & \textbf{3596} & 3596.0 & 47333 & \textbf{4895} & 4894.0 & 15877 \\
96 & \textbf{2484} & 2484.0 & 20119 & \textbf{3588} & 3587.4 & 26007 & \textbf{4898}* & 4897.2 & 56506 \\
97 & \textbf{2483} & 2482.2 & 12874 & \textbf{3596} & 3594.8 & 18748 & \textbf{4895} & 4894.6 & 52890 \\
98 & \textbf{2484} & 2483.6 & 17232 & \textbf{3595} & 3593.8 & 39620 & \textbf{4896} & 4895.8 & 34437 \\
99 & \textbf{2487} & 2487.0 & 11038 & \textbf{3596} & 3595.0 & 44668 & 4900* & 4900.0 & 48432 \\
100 & \textbf{2481} & 2481.0 & 6466 & \textbf{3592} & 3591.6 & 38165 & \textbf{4895} & 4895.0 & 41758 \\
 \hline
\end{tabular}
\label{table:detailed_PLSE_results_v1}
\end{table}
\renewcommand{\baselinestretch}{1}\large\normalsize




\section{Results on the Latin square completion problem}\label{sec:results_LSC}

The Latin square completion (LSC) problem can be considered as a special case of the partial Latin square extension problem. Two sets of LSC benchmark instances exist in the literature: 19 traditional instances \cite{gomes2002completing} and 1800 new instances \cite{haraguchi2016iterated}. These instances were built from complete Latin squares with some symbols removed. Thus these instances have the optimal score of $n^2$ ($n$ is the order of the grid), i.e., their cells can be completely filled. Like the 1800 PLSE benchmark instances, these 1800 LCS instances have an order $n \in \{50,60,70\}$ and ratio $r \in \{0.3,0.4,0.5,0.6,0.7,0.8\}$, grouped to 18 subsets of 100 instances per $(n,r)$ combination.

We ran the MPMA algorithm with a time limit of 3h with the parameters of Table \ref{table:parameters_MPMA} to solve the 1800 LCS instances. For the most difficult instances of the 19 traditional instances a time limit of 10 hours is required. The results on the set of 19 traditional instances (Table \ref{table:LSC_tradi_instances}) indicate that MPMA can solve all these instances with a perfect success rate. Only the best LSC algorithm MMCOL \cite{jin2019solving} reaches such a performance. However, MPMA requires a much higher computation time compared to the algorithm of \cite{jin2019solving}.

Table \ref{table:LSC_problem} displays the results of the MPMA algorithm on the set of 1800 LCS instances compared to the state-of-the-art algorithms \cite{haraguchi2016iterated,jin2019solving} presented in Section \ref{sec:benchmarks}. The results indicate that 
 MPMA is able to solve all of these 1800 instances in the allotted time, matching the best LSC algorithm of \cite{jin2019solving}.

 \renewcommand{\baselinestretch}{0.8}\large\normalsize

\begin{table}[!h]
\centering
\scriptsize
\caption{Results of the MPMA algorithm on the set of 19 traditional LSC instances \cite{gomes2002completing}.}
\begin{tabular}{lll|l|l} 
   \hline
 \multicolumn{3}{c|}{Instance}  &    \multicolumn{2}{c}{MPMA} \\
   \hline
   Name &  $n$ & $r$  & SR & t(s)    \\
    \hline
  qwhdec.order5.holes10.1 & 5 & 0.6 & 10/10 & 1.2 \\
  qwhdec.order18.holes120.1 & 18 & 0.63 & 10/10 & 1.9 \\  
  qg.order30 & 30 & 0.0  & 10/10 & 22  \\ 
  qwhdec.order30.holes316.1 & 30  & 0.65 & 10/10 & 12 \\ 
  qwhdec.order30.holes320.1 & 30  & 0.64 & 10/10 & 4  \\ 
  qg.order40 & 40 & 0.0 & 10/10 & 55 \\
  qg.order60 & 60  & 0.0 & 10/10 & 526   \\
  qg.order100 & 100  & 0.0 & 10/10 & 3864   \\  
 qwhdec.order35.holes405.1  & 35  & 0.67 & 10/10 & 56   \\ 
 qwhdec.order40.holes528.1 & 40 & 0.67 & 10/10 & 158  \\ 
 qwhdec.order60.holes1440.1 & 60  & 0.60 & 10/10 & 298  \\ 
 qwhdec.order60.holes1620.1 & 60 & 0.55 & 10/10 & 189  \\ 
qwhdec.order70.holes2940.1 & 70  & 0.4 & 10/10 & 546 \\
 qwhdec.order70.holes2450.1 & 70  & 0.5 & 10/10 & 356   \\  
  qwhdec.order33.holes381.bal.1 & 33 & 0.65 & 10/10 & 208  \\ 
 qwhdec.order50.holes825.bal.1 & 50 & 0.67 & 10/10 & 564   \\ 
  qwhdec.order50.holes750.bal.1 & 50 & 0.7 & 10/10 & 10546
  \\
qwhdec.order60.holes1080.bal.1  & 60 & 0.7 & 10/10 & 32484   \\
qwhdec.order60.holes1152.bal.1  & 60 & 0.68 & 10/10 & 9556   \\
\hline
\end{tabular}
\label{table:LSC_tradi_instances}
\end{table} 

 \renewcommand{\baselinestretch}{1.0}\large\normalsize

 \renewcommand{\baselinestretch}{0.8}\large\normalsize

\begin{table}[!h]
\centering
\scriptsize
\caption{Results of the MPMA algorithm on the 1800 new LSC instances \cite{haraguchi2016iterated} along with the results reported in the literature \cite{haraguchi2016iterated,jin2019solving}.}
\begin{tabular}{ll|l|l|l|l|l|l} 
   \hline
 \multicolumn{2}{c|}{Instance}  &    \multicolumn{1}{|c|}{CPX-IP} & CPX-CP & LSSOL  & Tr-ILS*  & MMCOL  &  \multicolumn{1}{|c}{MPMA} \\
   \hline
    $n$ & $r$ &  \#Solved & \#Solved & \#Solved & \#Solved & \#Solved  &  \#Solved   \\
    \hline
  \multirow{7}{*}{50} & 30  & 9 & 94 & 10   & 100 & 100  & 100  \\
  & 40  & 3 & 71 & 8 &  100 & 100 & 100 \\  
  & 50  & 0 & 12  & 6 &  100 & 100 & 100  \\ 
  & 60  & 0 & 0 & 0 & 36  & 100  & 100 \\ 
  & 70  & 0 & 0 &  0 & 0  & 100  & 100  \\ 
  & 80 & 100 &  100 & 100 & 100  & 100  & 100  \\
 \hline
  \multirow{6}{*}{60} & 0.3  & 0 & 71 & 1  & 100  & 100  & 100    \\
  & 0.4  & 0 & 22 & 0 &  100  & 100 & 100    \\  
  & 0.5  & 0 & 1 & 0 &  95 & 100 & 100   \\ 
  & 0.6  & 0 & 0 & 0 &  23 & 100  & 100  \\ 
  & 0.7 & 0 & 0 & 0  & 0 & 100  & 100   \\ 
  & 0.8 & 100 & 100 & 99  & 99  & 100 & 100  \\
   \hline
 \multirow{6}{*}{70} & 0.3  & 0 & 34 & 0  & 99   & 100  & 100   \\
  & 0.4  & 0 & 8  & 0  & 98  & 100 & 100   \\  
  & 0.5 & 0 & 0 & 0 &  84   & 100  & 100  \\ 
  & 0.6  & 0 & 0 & 0 &  10   & 100 & 100    \\ 
  & 0.7 & 0 & 0 &  0 & 0  & 100  & 100 
  \\
  & 0.8  & 100 & 100 & 46 &  98  & 100 & 100   \\
 \hline
\end{tabular}
\label{table:LSC_problem}
\end{table} 
 \renewcommand{\baselinestretch}{1.0}\large\normalsize

\end{document}